\documentclass[11pt]{article}
\usepackage[margin=1in]{geometry}
\usepackage{times}
\usepackage[round]{natbib}
\usepackage{authblk}

\usepackage[T1]{fontenc}
\usepackage[utf8]{inputenc}
\usepackage{amsmath,amssymb,booktabs,array,longtable,graphicx,xcolor}
\usepackage{enumitem}
\usepackage{hyperref}
\usepackage{url}
\definecolor{draftcolor}{RGB}{125,48,28}

\newcommand{\sfthree}{SceneFactory-3D}
\newcommand{\Lzero}{Frozen L0}
\newcommand{\Lzerot}{L0-T}
\newcommand{\Lone}{L1}
\newcommand{\muenv}{\mu_{\mathrm{surface}}}

\newcommand{\TierOneWorlds}{32}
\newcommand{\TierTwoWorlds}{1{,}024}
\newcommand{\TrainingSeeds}{4}

\newcommand{\DevBandFailureFrozen}{0.085}
\newcommand{\DevBandFailureTerrain}{0.025}
\newcommand{\DevBandFailurePreview}{0.024}

\newcommand{\LZeroOpenVPEighteen}{0.28}

\newcommand{\LOneDeltaEighteen}{$-$0.26}

\newcommand{\LOneDeltaEighteenTlo}{$-$0.59}
\newcommand{\LOneDeltaEighteenThi}{+0.06}

\newcommand{\LOneOpenVPEighteen}{0.62}

\newcommand{\BatchFlipsMax}{112}
\newcommand{\BatchFlipsMin}{70}
\newcommand{\BatchSameFlips}{2}

\newcommand{\ScaleVehStepsProd}{1.71}

\newcommand{\ScaleEffProd}{0.95}

\newcommand{\ScaleLoopSixtyFour}{27.0}
\newcommand{\ScaleLoopTenTwentyFour}{28.4}
\newcommand{\ScaleMemProd}{8.7}

\newcommand{\PDMLoopPerWorld}{30}
\newcommand{\PDMVehSteps}{1{,}565}

\newcommand{\LearnedLoopPerWorld}{0.032}
\newcommand{\LearnedLoopPerWorldSixtyFour}{0.42}
\newcommand{\PDMSlowdownSixtyFour}{72}
\newcommand{\PDMSlowdownProd}{961}
\newcommand{\IDMLoopPerWorld}{1.2}
\newcommand{\ComputeTierTwoHours}{12.2}
\newcommand{\ComputePDMHours}{5.9}
\newcommand{\ComputeIDMHours}{0.25}
\newcommand{\ComputeRAM}{125}
\newcommand{\ComputeThreads}{88}
\newcommand{\ComputeGPUMem}{96}

\title{SceneFactory-3D: Lifting 2D Traffic Scenes into 3D Physical Counterfactuals for Scalable Physically Grounded Safety Evaluation}
\author[1]{Yicheng Zhu}
\author[2]{Linfeng Tian}
\author[3]{Tianmu Zhao}
\author[1]{Yang Chen}
\author[3]{Fan Zuo}
\author[2]{Tao Li}
\author[1]{Zilin Bian\thanks{Corresponding author: \href{zxbite@rit.edu }{zxbite@rit.edu }}}
\affil[1]{Rochester Institute of Technology}
\affil[2]{City University of Hong Kong}
\affil[3]{New York University}
\date{}
\hypersetup{pdfauthor={Yicheng Zhu, Linfeng Tian, Tianmu Zhao, Yang Chen, Fan Zuo, Tao Li, Zilin Bian},pdftitle={SceneFactory-3D: Lifting 2D Traffic Scenes into 3D Physical Counterfactuals for Scalable Physically Grounded Safety Evaluation},hidelinks}

\begin{document}
\maketitle
\begin{abstract}
Scalable driving simulators typically execute vehicle commands using prescribed behavioral or kinematic rules, overlooking the physics of tire-road interfaces, thereby limiting their ability to capture how adverse road and environmental conditions alter vehicle execution and propagate through traffic. To address this limitation, we present SceneFactory-3D, a GPU-batched, physics-grounded multi-agent driving simulator. Vehicles execute acceleration and steering commands via suspension- and friction-limited forces evaluated at each wheel-contact point. Spatially varying friction, per-world 3D heightfields, gravity, and rigid contact consistently govern wheel motion and chassis collisions. Per-world terrain isolation and GPU batching enable SceneFactory-3D to run matched physical counterfactuals in parallel: traffic scenario setup and vehicle controllers remain fixed while only the road condition changes, enabling the resulting closed-loop effects to be evaluated across parallel worlds. To demonstrate the advantage of the \sfthree{}-enabled counterfactual evaluation, we conduct an empirical study on vehicle controllers' sensitivity to road conditions. We study three learned-policy families on 1,024 matched 12-vehicle worlds per condition, and two classical planners on a shared 32-world subset, across 21 friction and grade conditions. When the friction coefficient drops from 1 to 0.18 due to slippery conditions, the vehicle passage rate drops by 18\% and driving risk increases by 29\%.

\par\smallskip\noindent Code: \url{https://github.com/SmallWorldLab/SceneFactory_3D}
\end{abstract}
\section{Introduction}
Traffic simulation makes large interactive experiments tractable by skipping some physics-grounded modeling. For example, a car-following law or a prescribed kinematic rule scales thousands of
vehicle interactions at low cost~\citep{sumo,idm,waymax,gpudrive}. However, such rules also
fix how vehicle control execution responds to the physical environment: once a vehicle has decided to brake or
merge, the motion it realizes follows exactly the prescribed rule, which is detached from the underlying physics when the adverse road conditions affect vehicle control execution, resulting in a widening sim2real gap. 

To motivate the integration of vehicle-environment interacting physics with driving and traffic simulations, consider a temporary
lane closure planned for winter in a city such as Rochester. Although the proposed merge path may remain geometrically feasible on a low-friction road, vehicles’ ability to follow that path, e.g., to brake safely for one another, depends on the physical conditions at the tire–road interface. If one vehicle cannot complete the merge as intended, it may alter the behavior and trajectory of every vehicle behind it, potentially changing the traffic outcome. A credible evaluation, whether for self-driving testing or traffic management and planning, must therefore capture how the same scenario and controllers perform under different road conditions. Without this physical modeling, the simulation may substantially overestimate safety and traffic efficiency when adverse conditions affect vehicle dynamics.

While existing work accounts for physical friction in 2D planar driving simulations, comprehensive physics-grounded vehicle–environment interactions go beyond what these two-dimensional models can faithfully represent. Planar single-track models can enforce friction limits through tire-force constraints and friction-circle checks \citep{commonroad,commonroad_feasibility}, but they do not explicitly capture how \textbf{three-dimensional} road geometry and gravity affect wheel-level contact. Road grade, banking, and elevation change the normal force at each wheel and, consequently, the friction available to the vehicle. These effects become especially important when friction varies across the wheels of a vehicle or between lanes across diverse terrains. 

To support efficient and scalable simulation of many interacting vehicles across parallel 3D worlds with underlying physics,  we present \sfthree{}, a GPU-batched simulator that extends SceneFactory~\citep {scenefactory}, a GPU-vectorized Isaac Lab platform
whose articulated vehicles drive on flat roads with only a planar friction mechanism. \sfthree{} replaces the default vehicle model with a custom tire-force model evaluated at each wheel–road contact. It supports spatially varying friction patches and a separate terrain heightfield for each parallel world. Wheel-contact queries, terrain preview, and chassis collisions all reference the same heightfield, ensuring a consistent representation of the physical environment. Finally, a canonical-route interface provides classical and learned controllers with the same geometric route reference.

As a physics-grounded and GPU-batched 3D platform, \sfthree{} makes matched physical environment counterfactuals practical at scale. From the same traffic scene, initial state, routes, and controller population, it instantiates parallel counterfactual worlds that differ only in their physical road conditions. This design makes three factors directly controllable: where a physical condition occurs relative to a planned maneuver, whether the controller observes that condition, and how its effect on one vehicle propagates through interacting traffic. We demonstrate this capability using route-conditioned policies trained with PPO \citep{ppo}. We compare a policy trained on nominal roads with two continued-training variants exposed to diverse terrain as well as two classical planners, across 21 physical conditions. Although terrain randomization and privileged physical context are established techniques \citep{peng2018,rma,yu2017}, SceneFactory-3D enables their separation and evaluation via matched counterfactual worlds. Across 1,024 matched worlds per learned policy and condition, reduced friction degraded task completion for every controller, propagated to vehicles traveling in the unchanged lane, and was not anticipated by the preview-enabled policy. The magnitude of these effects varied more across training seeds than across training regimes, while the tested grades affected only a few policies.

In summary, our contributions are as follows. Section~\ref{sec:related} highlights our unique contributions by comparing \sfthree{} with prior efforts on driving simulation, co-simulation frameworks, and learning-based driving and vehicle controls using simulations.
\begin{enumerate}[leftmargin=0pt,topsep=0pt,itemsep=0pt]
    \item \sfthree{}: a \textbf{GPU-batched physics-grounded} multi-agent driving simulator where Per-wheel forces and rigid contact couple vehicle motion to spatially varying friction and 3D terrain. Per-world terrain ownership keeps wheel queries, observations, and collisions physically consistent.
    \item A matched \textbf{physical-counterfactual evaluation} framework. SceneFactory-3D creates matched counterfactual worlds that differ only in road conditions and 2D-to-3D geometry differences, enabling controlled comparisons across classical and learned vehicle controllers in terms of physical behaviors and safety risks. 
    \item An empirical study of vehicle controller \textbf{sensitivity to road physics}. Across 21 friction and grade conditions, we evaluate two classical planners and three learned-policy families that separate terrain-diverse training from explicit terrain preview. When the friction coefficient drops from 1 to 0.18 due to slippery conditions, the vehicle passage rate drops by 18\% and driving risk increases by up to 29\%.
\end{enumerate}

\begin{figure}[t]
\centering
\includegraphics[width=\linewidth]{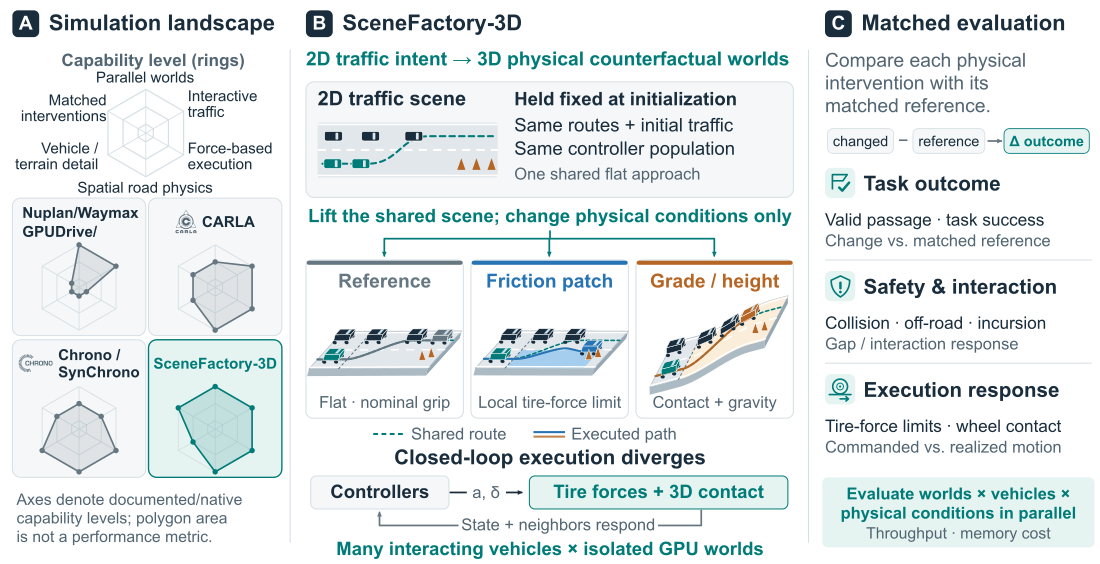}
\caption{\sfthree{} overview. (A)~Capability levels of representative
simulators on six axes (inner ring 0 to outer ring 2; criteria in
Appendix~\ref{app:spectrum}). \sfthree{} is the only system shown at the top level for
parallel worlds, interactive traffic, force-based execution, spatial road physics and
matched interventions; Chrono offers more vehicle and terrain detail. (B)~One planar
scene is instantiated as matched 3D worlds that differ only in road physics, so each
changed world is a counterfactual of the reference. (C)~Each counterfactual is
compared with its reference on task outcome, interaction safety and vehicle response.}
\label{fig:overview}
\vspace{-0.3cm}
\end{figure}

\section{Related Work}
\label{sec:related}
Traffic and driving simulators differ less in whether they recognize road
conditions than in which transition law executes a vehicle's commands.
Microscopic traffic simulators such as SUMO move vehicles with behavioral models
of car-following and lane changing~\citep{sumo}; the original Intelligent Driver
Model study represents road inhomogeneities such as uphill gradients through
modified driver parameters~\citep{idm}. Accelerator-based simulators extend this
trade to learning at scale: Waymax advances agents with kinematic bicycle or
position-delta dynamics, and GPUDrive by default with a kinematic bicycle model,
over compact state~\citep{waymax,gpudrive}; neither has tire forces or a friction
limit. Planar dynamic models occupy the middle ground. CommonRoad provides
point-mass, kinematic and dynamic single-track and multi-body vehicle models, and
checks trajectories against a friction circle~\citep{commonroad,commonroad_feasibility}.
Every simulator takes friction as an input; the relevant distinction is between
prescribing a behavioral response to the road and computing execution through an
explicit force and contact model.

Driving environments built on physics engines compute execution in this way,
with different arrangements of the surrounding traffic. CARLA simulates vehicles
in a game-engine physics model with localized friction triggers and an optional
hybrid mode, off by default, that teleports distant background
vehicles~\citep{carla,carla_friction,carla_tm}; SUMMIT extends it with a
velocity-space model of massive mixed traffic~\citep{summit}; SMARTS combines
Bullet-simulated vehicles of interest with SUMO background traffic~\citep{smarts};
and MetaDrive composes diverse scenarios on a physics simulator~\citep{metadrive}.
Chrono::Vehicle offers detailed subsystems, several tire models and deformable
terrain~\citep{chrono,chrono2016}, and SynChrono distributes interacting agents across
compute nodes~\citep{synchrono}, an architecture different from batching many
worlds in one accelerator process. Figure~\ref{fig:overview}A and
Appendix~\ref{app:spectrum} compare the capabilities we could verify.

Several recent systems are closer to our motivation. SceneFactory batches
articulated PhysX vehicles on flattened roads converted from the Waymo Open
Motion Dataset, maps weather to one friction coefficient per world, and compares
policies with and without a weather-context input~\citep{scenefactory}; its own
limitations name flat geometry and spatially uniform friction. Its two policies
differed both in wet-road training exposure and in the context input, whereas
our learned arms separate these factors. \sfthree{}
inherits its GPU-batched, Isaac Lab-based design and replaces the vehicle, road
and evaluation layers (Appendix Table~\ref{tab:predecessor}).
$\mu$Sync-GS couples friction and road-elevation edits to the response of a
single ego vehicle, modeled as a half-car with a Burckhardt tire curve, to
synthesize consistent driving video~\citep{musyncgs}; it involves no traffic
interaction, and \sfthree{} shares neither its vehicle model nor its rendering.
MultiDrive co-simulates CommonRoad scenarios in the BeamNG.tech simulator and
reports discrepancies between a motion planner's intended and executed
behavior~\citep{multidrive}. Converting scenarios between simulators, executing
them with richer physics, and comparing intended with realized motion are thus
established. ScenarioNet likewise converts real-world driving logs into
interactive simulation scenarios~\citep{scenarionet}. Our emphasis is a batched
workflow in which physical interventions are authored, matched and repeated
across many interacting vehicles and learned controllers.

On the learning side, training under randomized dynamics~\citep{peng2018} and
conditioning a policy on physical parameters, either given or
inferred~\citep{yu2017,rma,tao23cola}, are established tools; contextual reinforcement
learning formalizes when such context must be observed~\citep{carl,martinez2026context}. Friction
preview has also been used explicitly in model-predictive trajectory
tracking~\citep{gao2022}. Our terrain-trained arm without preview corresponds to
randomization with latent context, and our preview arm to privileged explicit
context, both for a feed-forward policy. What the simulator adds is control over
the spatial placement of the physical context, over the interacting population
that experiences it, and over matched execution of the same scenario. Closed-loop
planning benchmarks such as nuPlan evaluate a planner by the traffic outcomes its
decisions produce~\citep{nuplan}; our experiments hold the scenario and controller
population fixed while the physical world changes.

\section{Problem Setting}
\label{sec:problem}
A controlled physical-execution experiment separates three inputs that a
simulator normally fuses: the scenario, the controllers, and the physical road.
A scenario $q$ comprises the planar road layout, a route $r_i$ for each of the
$M$ vehicles, the vehicle models, and the initial state $x_0$. The initial state
includes chassis pose and velocity, wheel spin (set to rolling without slip),
steering and suspension state, each controller's held command and internal
state, and each vehicle's progress along its route. A physical condition $p$
specifies the road's height field and friction patches. A controller assignment
$\pi=(\pi_1,\ldots,\pi_M)$ gives the policy of every vehicle; in our benchmark
all $M$ vehicles run the same controller with shared parameters. A rollout is
\begin{equation}
 u_{i,t}=\pi_i\!\left(o_i(x_t,q,p)\right),\qquad
 x_{t+1}=F_{\theta,p}(x_t,u_t),\qquad
 \tau(q,\pi,p)=(x_0,u_0,x_1,\ldots),
 \label{eq:rollout}
\end{equation}
where $\theta$ holds fixed vehicle and numerical parameters. The physical
condition enters in two places: it always shapes the transition $F_{\theta,p}$,
and it enters the observation $o_i$ only for controllers given physical inputs.
In our evaluation both the transition and every controller are deterministic
(learned policies act with the mean of their action distribution), so a rollout
is a function of $(q,\pi,p)$ up to floating-point effects.

For an outcome $M$ and reference condition $p_0$, the physical contrast is
\begin{equation}
 \Delta_p M(q,\pi)=M\!\left(\tau(q,\pi,p)\right)
                    -M\!\left(\tau(q,\pi,p_0)\right).
 \label{eq:contrast}
\end{equation}
It is the total closed-loop effect of the intervention within the simulator.
Only $q$ and $\pi$ are held fixed. Commands and trajectories diverge afterward:
a vehicle that slides on a low-grip patch changes what its own controller and its
neighbors do next, and, for a controller with physical inputs, the changed
observation adds a second pathway. Averaging $\Delta_p M$ over scenarios drawn
from a generator gives a paired comparison of the kind that common random
numbers make efficient in stochastic simulation~\citep{glasserman1992}; pairing
is not by itself a new causal method, and interpreting a contrast for a real road
additionally requires validating the model. Comparing controller assignments
instead fixes $p$ and varies $\pi$. Because every vehicle switches controller
together, that comparison is between population regimes, not an estimate of one
vehicle's skill against fixed surrounding traffic.

Equation~\ref{eq:contrast} requires only identical $(q,\pi)$ on both sides and a
transition that depends on $p$; it can be evaluated sequentially or with a
lower-dimensional model. Our implementation choices serve the questions of
Section~\ref{sec:related} and the evaluation budget: the road acts through
per-wheel forces and contact; routes are defined without reference to any
controller; batched worlds evolve under their own $p$ without interacting; and
the product of scenes, conditions, controllers and seeds must fit on one
accelerator. Section~\ref{sec:system} describes each.

\section{SceneFactory-3D}
\label{sec:system}
\sfthree{} implements the design choices of Section~\ref{sec:problem} with four
mechanisms, described in the order a rollout uses them: canonical routes give
every controller the same reference; a force-based vehicle model makes the road
part of the dynamics; per-world terrain ownership keeps paired worlds separate;
and GPU batching makes the combinatorial evaluation practical.

\subsection{Canonical routes}
A canonical route is a planar path planned once from the road layout, without
reference to any controller: $r(u)=c(u)+o(u)\,n(u)$, where $c$ is the road
reference line, $n$ its normal, and $o$ a lateral offset profile. Lane following
uses a constant offset; a merge blends two lane offsets with the quintic
$10t^3-15t^4+6t^5$, which has zero curvature at both ends. Routes are resampled
at 0.5~m of arc length and checked for continuity, drivable bounds, 1.2~m
clearance from the closed lane, and curvature below 0.0139~m$^{-1}$, a budget
derived from the vehicle's speed-dependent steering limit (Appendix~\ref{app:controllers}).
These checks concern the path alone. Whether it can be executed at a given speed
and friction level, or by several vehicles at once, is what the experiment
measures.

Every controller receives its route, route-relative samples ahead of the vehicle
and the current traffic state, and returns a longitudinal acceleration and a
normalized steering command. The route is therefore a shared spatial reference,
not a shared trajectory: controllers choose speed, gaps and timing, and the
proposal-based planner may track lateral offsets of $\pm1$~m around it. No route
deviation invalidates a vehicle by itself; validity is defined by the outcome
rules of Section~\ref{sec:experiments}. Route arc length, curvature and preview
distances are planar quantities, and on sloped terrain the vehicle's position is
projected onto the plane.

\subsection{Vehicle and road model}
\label{sec:vehicle}
Each vehicle is a free rigid chassis with four raycast suspension legs and
rotating wheels. Newton integrates the chassis and resolves chassis contacts;
a custom Warp kernel computes, at every 240~Hz step, suspension, tire, drive,
brake, steering and drag forces and applies them to the
chassis~\citep{newton,warp2022}. At wheel $j$, a ray along the chassis down axis
finds the road; the spring--damper load is $F_{z,j}=\max(0,\,k\delta_j-c\,\dot\delta_j)$
plus a stiffer bump stop, and the contact frame is aligned with the surface
normal. With contact-frame slip velocities $(v_x,v_y)$, wheel speed $\omega$ and
radius $R$, the tire forces are
\begin{equation}
\begin{gathered}
 F_x=k_x\,(R\tilde\omega-v_x),\qquad
 F_y=-C_\alpha\operatorname{atan2}\!\left(v_y,\max(|v_x|,2)\right),\\
 \|(F_x,F_y)\|\le\mu_{\mathrm{vehicle}}\,\muenv(c_j)\,F_{z,j},
\end{gathered}
 \label{eq:friction}
\end{equation}
where $k_x=70{,}000/\max(|v_x|,2)$~N\,s/m regularizes the longitudinal slip
stiffness, $\tilde\omega$ is the wheel speed advanced implicitly with drive and
brake torques, and $C_\alpha=55{,}000$~N/rad. When the combined force exceeds the
bound, both components are scaled onto it; rolling resistance shares the same
budget, and the wheel speed is updated with the resulting reaction torque. This
is a linear-slip, arctangent-slip-angle approximation with friction-circle
saturation, not a Magic Formula or brush model~\citep{pacejka2012}; the vehicle
grip factor is $\mu_{\mathrm{vehicle}}=1.05$. Commands pass through one actuation
bridge shared by all controllers: a requested acceleration $a$ becomes throttle
$\min(1,\max(0,a/4.91))$, where 4.91~m/s$^2$ is the drive-torque limit divided by
$Rm$, or brake $\min(1,\max(0,-a/8))$, and normalized steering sets a
rate-limited target angle that shrinks with speed. Requested acceleration is
therefore not the realized acceleration, and a requested deceleration or turn is
realized only as far as the road's grip allows. Appendix~\ref{app:vehicle} lists
all parameters, clamps and the integration sequence.

Road conditions are authored as per-world heightfields and friction patches
(discs or rotated rectangles, with later patches taking precedence). Friction
changes the bound in Equation~\ref{eq:friction} at the contact point of each
wheel; height changes contact geometry and the component of gravity along the
road. Condition parameters describe surface grip and grade; they are not a
weather model. Appendix~\ref{app:validation} verifies that each benchmark
condition reaches the wheel contacts.

\subsection{World isolation and matching}
A matched experiment runs the reference and changed conditions, and many
scenes of each, side by side. All worlds share one batched simulation model with
world-indexed bodies and collision shapes; collisions are resolved within a
world and never between worlds. The road must also mean the same thing to every
physical subsystem of a world: wheel rays, route and preview queries, and
chassis collision geometry all resolve the terrain variant assigned to that
world; updating
wheel queries while chassis colliders kept a shared terrain would silently mix
two physical worlds, a failure mode the terrain checks exclude (Appendix~\ref{app:validation}).
Worlds that use the same variant share its geometry. Terrain is fixed when an
environment is constructed and retained across resets, and a partial reset
restores the dynamic state of the affected world alone.

Each run records scenario and route digests, the physical initial state, the
terrain hash, controller and checkpoint identity, and timing settings. Before a
comparison is admitted, a matching check requires the paired runs to differ only
in the fields of the declared intervention axis: terrain for a physical
 contrast, controller for a controller contrast.
% Matching is established at
% initialization only: the interface does not branch a running traffic state, and
% trajectories then diverge through feedback and interaction, as
% Equation~\ref{eq:contrast} intends.

\subsection{GPU batching}
The benchmark in Section~\ref{sec:experiments} comprises about 259{,}000
twelve-vehicle world rollouts, so \sfthree{} steps all worlds together on the
GPU: an Isaac Lab \texttt{DirectRLEnv}~\citep{isaaclab2025} runs Newton's XPBD
solver~\citep{isaaclab_backends,newton,xpbd} without rendering, with CUDA graphs,
240~Hz physics and 60~Hz actuation; controllers may decide less often and hold
their commands. Scene construction and some controllers run on the CPU. Existing
checks cover up to 64 distinct terrains in one batch; the throughput and memory
study of Section~\ref{sec:results} measures the cost of scale.

\section{Experimental Setup}
\label{sec:experiments}
\begin{figure}[t]
\centering
\includegraphics[width=\linewidth]{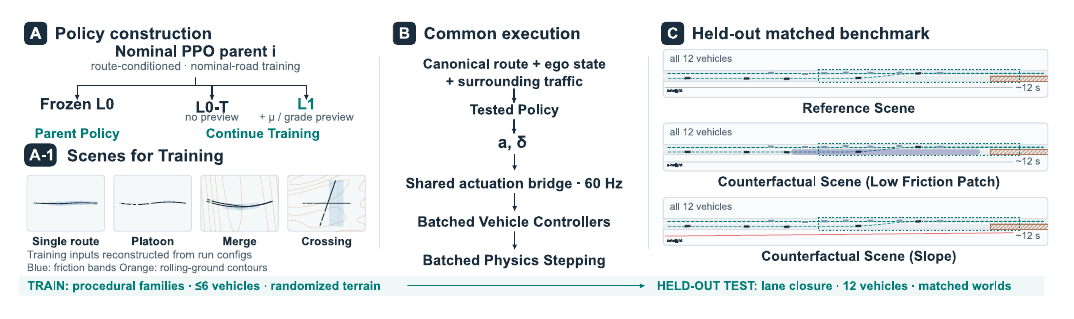}
\caption{Controller construction and matched evaluation. (A)~Each of four
nominal PPO parents is kept frozen (\Lzero{}) or continued on terrain-diverse training
scenes without (\Lzerot{}) or with (\Lone{}) a friction and grade preview. (B)~All
controllers share route and traffic inputs, the actuation bridge and batched physics.
(C)~The held-out closure task runs on a reference road and on matched counterfactual
roads with a low-friction patch, a slope, or combined.}
\label{fig:pipeline}
\vspace{-0.2cm}
\end{figure}

The benchmark instantiates the contrast of Equation~\ref{eq:contrast} with one
driving task, one set of canonical routes, five controller regimes and a matrix
of 21 physical conditions, evaluated on matched worlds (Figure~\ref{fig:pipeline}).

\noindent\textbf{Task and routes.}
Each world is a straight two-lane road whose right lane closes at
$x=150$~m. Of its 12 vehicles, five start in the closing lane and follow a route
that merges into the open lane between 50 and 110~m; seven follow the open lane.
Worlds differ in initial positions and speeds, the speed of the open-lane stream
(2--8~m/s), the presence of a closing-lane leader, and per-vehicle driver
presets. The presets set headway, minimum gap and acceleration limits for
IDM-17 only; the learned policies and PDM-Closed-17 do not read them. In every
world, all 12 vehicles are driven by the evaluated controller with shared
parameters and execute through the physical vehicle model, so a vehicle that
slides or brakes late changes the situation of its neighbors through the same
dynamics. Episodes last 66~s.

\noindent\textbf{Controller regimes.}
Figure~\ref{fig:pipeline}B and Table~\ref{tab:controllers} (Appendix~\ref{app:controllers}) list the five regimes. Of the two classical
controllers, IDM-17 combines the IDM longitudinal law~\citep{idm}
with Stanley-style route tracking~\citep{stanley}, and PDM-Closed-17 adapts the
published PDM-Closed planner~\citep{pdm} to the canonical route and drivable
road. Both use a 17~m/s target speed to reduce differences
in operating speed. The three learned regimes share one route-conditioned
feed-forward policy architecture trained with PPO~\citep{ppo}, observing the ego
vehicle, its route and the surrounding traffic, with \TrainingSeeds{} training
seeds each. They form two contrasts: \Lzero{} versus \Lzerot{} compares
nominal-road training with continued training on terrain-diverse roads, both
without physical observations, and \Lzerot{} versus \Lone{} compares the same
terrain-diverse continuation without and with a route-aligned preview of surface
friction and grade at six distances from 0 to 40~m ahead.
Each \Lzerot{} and \Lone{} seed continues from the \Lzero{} checkpoint with the
same seed, so both contrasts are paired by seed. The second contrast is matched
in training. The first is not: the terrain-trained arms also receive additional
training, so \Lzero{} versus \Lzerot{} measures the combined effect of the
continuation regime until an equal-budget nominal continuation is available.
\Lzerot{} must infer the road condition from its own motion, whereas \Lone{}
receives it as privileged context; neither has memory, so neither represents
history-based adaptive control. Training used different scene families, at most
six vehicles per world, friction bands up to 70~m long with $\muenv\in[0.2,0.7]$
or whole-road friction in $[0.35,0.8]$, and rolling rather than constant grades; the benchmark therefore tests unseen
geometry and, below $\muenv=0.2$, friction extrapolation
(Appendix~\ref{app:controllers}).

\noindent\textbf{Physical conditions.}
All conditions share a flat reference road up to 40~m, so every world begins
under identical physics. Friction conditions lower the surface coefficient on
the closing lane between 40 and 180~m, which contains the merge; grade
conditions tilt the full road width from the same point, with smooth
transitions into and out of a constant-grade section. The 21 conditions
are the reference road, eight friction levels ($\muenv$ from 0.70 to 0.05),
eight grades ($+4\%$ to $-8\%$) and four grade--friction compositions
(Appendix~\ref{app:conditions}). Both tiers use the complete matrix, fixed in a
freeze manifest with code, checkpoint and terrain hashes before production.
In a controller-free probe, a vehicle that brakes fully on entering the
140~m patch at 16~m/s stops within it for $\muenv\geq0.12$ but not for
$\muenv\in\{0.08,0.05\}$ (Appendix~\ref{app:validation}). This is a stopping
limit for one entry state, not a task limit: a controller can slow down before
the patch or complete its merge onto the dry open lane. We mark these two levels
as an extreme-severity group and discuss them separately.

\noindent\textbf{Evaluation tiers.}
Tier~1 evaluates all five regimes on \TierOneWorlds{} held-out worlds for a
common comparison. Tier~2 evaluates the three learned regimes on
\TierTwoWorlds{} worlds, whose first \TierOneWorlds{} are the Tier-1 worlds;
the per-vehicle cost of the published PDM-Closed planner motivates the smaller
all-controller tier. Every controller is compared with its own reference rollouts
on the same worlds. All runs used one workstation GPU (Appendix~\ref{app:vehicle}).

\noindent\textbf{Outcomes.}
A vehicle is valid until its first closure incursion (body center in the closed
lane beyond $x=150$~m), contact with another vehicle, off-road departure (body
center outside the paved road) or environment fault; validity never recovers.
The primary task outcome, valid passage, requires the front bumper to cross a
counting line at $x=170$~m while the vehicle is still valid. Vehicles keep
driving and interacting after they cross. Strict task success additionally
requires the owed merge and validity over the whole horizon. Collision, off-road
and incursion rates are reported separately as safety and constraint events;
collision alone is insufficient as a safety measure because a vehicle that fails
early has less opportunity to collide. If any vehicle leaves the finite road mesh
or overturns, its world resets; nothing after the first reset is scored, and
valid vehicles that have not crossed by then count as censored non-passes in a
fixed denominator of 12 vehicles. Appendix~\ref{app:conditions} gives the exact
rules, including a map-edge adjustment reported beside the strict measure. We
call a controller's change in valid passage relative to its own reference
condition its \emph{sensitivity} to the intervention. Sensitivity is not a
quality ranking: it is always reported with the absolute reference and
changed-condition passage, since a controller that fails on both has none to
lose.

\noindent\textbf{Statistics.}
The world is the statistical unit, paired across conditions and controllers by
scenario seed. Finite-scene uncertainty for a fixed checkpoint and variability
across the four training seeds are reported separately
(Appendix~\ref{app:conditions}). No cell was excluded or rerun; all quantities use
each world's first episode (Appendix~\ref{app:final}).

\section{Results}
\label{sec:results}
Probes verify that all conditions reach the dynamics (Appendix~\ref{app:validation});
we report their closed-loop effect with paired world-bootstrap intervals.

\begin{table}[tb]
\centering\scriptsize
\setlength{\tabcolsep}{2.4pt}
\caption{Valid passage on the reference road and its change ($\Delta$) on the same
worlds at $\muenv=0.35$, $0.18$, $\{0.08,0.05\}$ (mean) and the most influential
grade ($\downarrow$: decrease); TTC: share of vehicles below 1.5~s (reference).
Tier~2: seed mean (range over seeds).}
\label{tab:final}
\resizebox{\linewidth}{!}{\begin{tabular}{@{}lcccccccc@{}}
\toprule
 & & \multicolumn{5}{c}{Valid passage} & $P(\mathrm{TTC}$ & Speed \\
\cmidrule(lr){3-7}
Regime & Worlds & Ref. & $\Delta\,\mu{=}.35$ & $\Delta\,\mu{=}.18$ & $\Delta\,\mu{\le}.08$ & $\Delta$ grade & ${<}1.5\,\mathrm{s})$ & (m/s) \\
\midrule
\multicolumn{9}{@{}l}{\emph{Tier 2: 1{,}024 worlds per checkpoint; mean over 4 training seeds (min, max over seeds)}}\\
Frozen L0 & 1{,}024 & .81\,{\tiny(.28, .99)} & $\downarrow$.27\,{\tiny($\downarrow$.70, $\downarrow$.01)} & $\downarrow$.57\,{\tiny($\downarrow$.90, $\downarrow$.21)} & $\downarrow$.69\,{\tiny($\downarrow$.93, $\downarrow$.12)} & $\downarrow$.05\,{\tiny($\downarrow$.10, +.00)} & .06\,{\tiny(.01, .14)} & 16.3\,{\tiny(14.7, 17.5)} \\
L0-T & 1{,}024 & .87\,{\tiny(.53, 1.00)} & $\downarrow$.12\,{\tiny($\downarrow$.36, $\downarrow$.00)} & $\downarrow$.49\,{\tiny($\downarrow$.83, $\downarrow$.07)} & $\downarrow$.70\,{\tiny($\downarrow$.89, $\downarrow$.42)} & $\downarrow$.03\,{\tiny($\downarrow$.13, +.00)} & .15\,{\tiny(.01, .51)} & 14.9\,{\tiny(12.0, 20.1)} \\
L1 & 1{,}024 & .86\,{\tiny(.48, 1.00)} & $\downarrow$.05\,{\tiny($\downarrow$.18, $\downarrow$.00)} & $\downarrow$.26\,{\tiny($\downarrow$.47, $\downarrow$.06)} & $\downarrow$.65\,{\tiny($\downarrow$.93, $\downarrow$.35)} & $\downarrow$.06\,{\tiny($\downarrow$.15, +.00)} & .07\,{\tiny(.01, .20)} & 14.8\,{\tiny(11.0, 19.9)} \\
\midrule
\multicolumn{9}{@{}l}{\emph{Tier 1: classical controllers on the first 32 Tier-2 worlds (learned rows: Table~\ref{tab:tier1})}}\\
IDM-17 & 32 & .68 & $\downarrow$.01 & $\downarrow$.18 & $\downarrow$.46 & $\downarrow$.08 & .46 & 11.6 \\
PDM-Closed-17 & 32 & 1.00 & $\downarrow$.06 & $\downarrow$.19 & $\downarrow$.47 & $\downarrow$.04 & .01 & 11.3 \\
\bottomrule
\end{tabular}
}
\end{table}

\noindent\textbf{Reference behavior.} On the reference road nine of twelve learned checkpoints pass
0.96--1.00 of vehicles (Table~\ref{tab:final}); the three seed-3 checkpoints (one parent)
pass 0.28--0.53 because in nearly every world a vehicle weaving at 27--33~m/s rolls
over and the reset censors the rest. Batch size
perturbs individual worlds (\BatchFlipsMin--\BatchFlipsMax{} of 768 vehicle outcomes
changed; reruns: at most \BatchSameFlips{}) without shifting mean passage, so
classical (32-world batches) versus learned Tier-1 differences are read only when large.

\begin{figure}[tb]
\centering
\includegraphics[width=\linewidth]{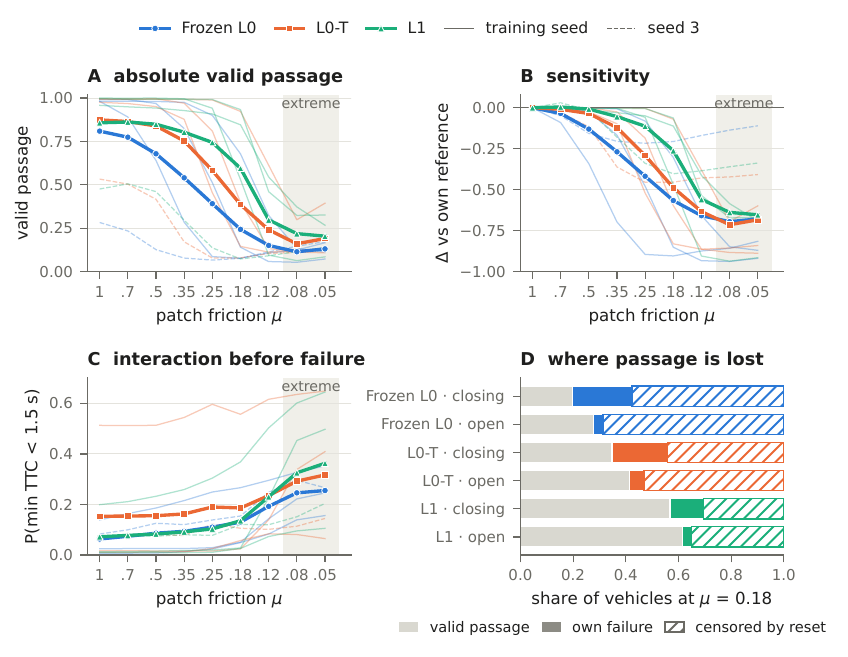}
\caption{Friction sweep on 1{,}024 matched worlds per policy. Thick lines:
family means; thin: seeds (dashed: seed~3); shaded: extreme levels. (A)~Valid passage.
(B)~Change from each policy's own reference. (C)~Share of vehicles with TTC below
1.5~s before their first failure. (D)~Where passage is lost at $\muenv=0.18$, by lane.}
\label{fig:robustness}
\vspace{-0.5cm}
\end{figure}

\noindent\textbf{Friction and grade.} Reduced friction lowered valid passage for every checkpoint by amounts the vehicle
equations do not determine: 0.06--0.90 at $\muenv=0.18$ (every interval excluding
zero) and 0.00--0.70 at $\muenv=0.35$ (Figure~\ref{fig:robustness}A,B). Training
seed mattered more than family: seed ranges within a family at $\muenv=0.18$
(0.41--0.76) exceed the differences between family means, and four-seed
$t$-intervals include zero for every family (e.g.\ \Lone{} \LOneDeltaEighteen{}
[\LOneDeltaEighteenTlo, \LOneDeltaEighteenThi]). The patch covers only the closing
lane, yet open-lane passage fell from 0.82--0.88 to
\LZeroOpenVPEighteen--\LOneOpenVPEighteen{} (Figure~\ref{fig:robustness}D). Open-lane
collisions rose only from about 1\% to 3--6\%; most of the loss is censoring when a
sliding closing-lane vehicle overturns and resets its world. At
$\muenv\le0.08$ the spread between checkpoints shrinks from 0.90 to 0.32, and passage
turns non-monotone through fewer resets, not better closing-lane driving. Grade changed passage by at most 0.011 for seven of twelve checkpoints; the largest
loss (0.15 at $-8\%$, \Lone{} seed~0) came with collisions rising from 4.9\% to 22.1\% (Figure~\ref{fig:grade}). Grade--friction compositions do not depart
resolvably from additivity (Appendix~\ref{app:final}).

\noindent\textbf{Interaction safety.} Within the valid-driving window, the share of vehicles reaching TTC below 1.5~s
rose for all twelve checkpoints at every $\muenv\le0.12$ (36 of 36 intervals exclude
zero; Figure~\ref{fig:robustness}C), although valid driving per vehicle fell from
about 53~s to 16--25~s; time below 1.5~s per second of valid driving rose for all
twelve already at $\muenv=0.18$. \Lzerot{} seed~0 brings 51\% of its vehicles below
1.5~s on the reference road but passes 98\%. In the 60~Hz subset, braking need above the capacity left after lateral
tire use 1--2~s earlier raised the hazard of a vehicle's own failure 11- to 30-fold
at $\muenv\le0.12$, and 10- to 25-fold while TTC was still above 1.5~s
(Table~\ref{tab:hazard}); ignoring timing reverses this sign (immortal-time
bias). These are associations, not causal effects.

\noindent\textbf{Training exposure and preview.} \Lone{} and \Lzerot{} differ only in the preview. \Lone{} lost less passage at
$\muenv=0.35$--$0.18$ in three of four seeds (world intervals excluding zero), but
mostly through seed~0 ($+0.72$ at 0.18, against $+0.13$ for seed~1 and near zero for
seed~2 at ceiling), with mixed signs at $\muenv\le0.12$ and a small reference-road
cost in two seeds; across seeds the contrast is unresolved (Table~\ref{tab:contrasts}).
The preview produced no anticipation of low friction: speed 20~m before the patch
changed by $-0.2$ to $+1.3$~m/s for every checkpoint. \Lone{} did anticipate grade,
entering a $-8\%$ slope 1.0--2.4~m/s slower.
Slower checkpoints lost less (Spearman $\rho=-0.83$ without seed~3, $-0.46$ with
it), yet \Lone{} seed~1 drove 3.6~m/s faster than \Lzerot{} seed~1 and lost less.
\Lzerot{} lost less than \Lzero{} in three of four seeds, with additional training.

\noindent\textbf{Classical and learned regimes.} Tier~1 compares population regimes that differ in planner, decision rate and
speed. PDM-Closed-17
passed every vehicle on the reference road, lost 0.06 at $\muenv=0.35$ and 0.19 at
$0.18$, and still passed 0.52--0.54 at $\muenv\le0.08$, at least 0.13 more than any
other regime (Table~\ref{tab:tier1}). IDM-17 lost 0.18 at $0.18$ from a lower
reference (0.68). Both classical losses at $\muenv=0.18$ lie inside the learned
range ($-0.02$ to $-0.91$). PDM-Closed-17 is also far costlier:
\PDMLoopPerWorld~s of loop time per world on 48 CPU workers, mostly planning, against
\LearnedLoopPerWorldSixtyFour~s for a learned policy (Appendix~\ref{app:scaling}).

\noindent\textbf{Computational scaling.} On one GPU, loop time grew from
\ScaleLoopSixtyFour{} to \ScaleLoopTenTwentyFour~s between 64 and 1{,}024
twelve-vehicle worlds (efficiency \ScaleEffProd), reaching \ScaleVehStepsProd{}
million vehicle-steps/s with \ScaleMemProd~GB peak memory, without saturating
(Appendix~\ref{app:scaling}).

\noindent\textbf{Discussion and Limitations}
The vehicle equations fix how much braking a patch removes, not how much a
controller population loses; that loss depended more on training seed than on
regime, so one policy per method could have produced any ordering. The simulator is
verified but not calibrated (Appendix~\ref{app:vehicle}), and simulator safety
measures are not evidence about real-road safety. The study uses one closure
geometry, one policy architecture and noise-free preview; four seeds cannot resolve
family differences of the observed size, and \Lzero{} versus \Lzerot{} includes
additional training. Whole-world resets couple vehicle outcomes, and batch
composition perturbs individual trajectories.

\section{Conclusion}

We presented SceneFactory-3D, a GPU-batched multi-agent driving simulator that couples vehicle motion to spatially varying friction and 3D terrain through per-wheel forces, suspension, gravity, and rigid contact. Per-world terrain isolation, canonical routes, and GPU batching enable matched physical counterfactuals in which road conditions change while the traffic scene, initial state, vehicles, and controllers remain fixed.
Across 21 friction and grade conditions, reduced friction degraded task completion for every controller, increased interaction risk, and affected vehicles outside the altered lane.  The throughput scaled nearly linearly to 1,024 worlds on one GPU. Overall, the results show that a geometrically valid driving policy may produce substantially different traffic outcomes when physical conditions change. 

\label{maintext:end}
\clearpage

\bibliography{references}
\bibliographystyle{iclr2027_conference}
\clearpage
\appendix
\section{Abstraction and Intended Use}
\label{app:spectrum}
The design space concerns modeling choices, not a measured ranking.
Table~\ref{tab:capabilities} records how each system executes vehicle motion
and surrounding traffic, and whether it supports spatially local friction,
non-flat roads and parallel execution, using only the cited sources. ``n.v.''
means not verified from those sources, not absent; entries describe native or
documented configurations, and many systems can be extended by custom
integration. A model can be sufficient for one outcome and insufficient for
another, and the relevant question for \sfthree{} is whether physical execution
alters the conclusion of a multi-agent experiment.

Panel~A of Figure~\ref{fig:overview} scores four systems on six ordinal axes
from each system's paper or official documentation. \emph{Parallel worlds}:
serial only (0), separate instances or distributed nodes (1), many worlds batched
in one accelerator process (2). \emph{Interactive traffic}: one vehicle (0),
several vehicles without native per-step mutual reaction (1), native closed-loop
multi-agent control (2). \emph{Force-based execution}: kinematic or behavioral
(0), planar force-limited (1), 3D wheel-level tire and contact forces (2).
\emph{Spatial road physics}: neither (0), either (1) or both (2) of local friction
and non-flat roads. \emph{Vehicle/terrain detail}: bicycle abstraction (0), rigid
chassis with wheel-level approximation (1), multibody subsystems, several tire
classes or deformable terrain (2). \emph{Matched interventions}: no road-physical
parameter (0), scriptable condition changes with user-built pairing (1), native
matched-world workflow (2).

\begin{table}[htb]
\centering\small
\caption{Capabilities as documented in the sources cited in Section~\ref{sec:related}
and this appendix~\citep{sumo,waymax,gpudrive,commonroad_feasibility,carla,carla_tm,smarts,summit,metadrive,chrono,synchrono,multidrive,scenefactory}.
n.v.: not verified here.}
\label{tab:capabilities}
\footnotesize\setlength{\tabcolsep}{3pt}
\begin{tabular}{@{}>{\raggedright\arraybackslash\hspace{0pt}}p{0.16\linewidth}>{\raggedright\arraybackslash\hspace{0pt}}p{0.21\linewidth}>{\raggedright\arraybackslash\hspace{0pt}}p{0.2\linewidth}>{\raggedright\arraybackslash\hspace{0pt}}p{0.12\linewidth}>{\raggedright\arraybackslash\hspace{0pt}}p{0.1\linewidth}>{\raggedright\arraybackslash\hspace{0pt}}p{0.12\linewidth}@{}}
\toprule
System & Vehicle execution & Surrounding traffic & Local friction & Non-flat road & Parallel execution \\
\midrule
SUMO & behavioral models & same & n.v. & n.v. & n.v. \\
Waymax & kinematic bicycle or position delta & same & no tire model & n.v. & accelerator \\
GPUDrive & kinematic bicycle (default) & same & no tire model & n.v. & GPU, $>10^6$ steps/s \\
CommonRoad & planar models incl.\ dynamic single-track; friction-circle checks & scenario agents & n.v. & n.v. & n.v. \\
CARLA & engine vehicle physics & physics; optional hybrid mode teleports distant vehicles & friction triggers & n.v. & n.v. \\
SMARTS & Bullet for vehicles of interest & SUMO & n.v. & n.v. & n.v. \\
SUMMIT & CARLA-based & velocity-space crowd model & n.v. & n.v. & n.v. \\
MetaDrive & physics simulation & n.v. & n.v. & n.v. & n.v. \\
Chrono & detailed subsystems, several tire models & SynChrono agents on separate nodes & n.v. & deformable terrain & distributed nodes \\
MultiDrive & BeamNG.tech co-simulation of CommonRoad scenarios & n.v. & n.v. & n.v. & n.v. \\
SceneFactory & articulated PhysX vehicles & same & one value per world & flattened & Isaac Lab worlds \\
\sfthree{} & per-wheel tire and contact forces & same & per-contact patches & heightfields & Isaac Lab / Newton worlds \\
\bottomrule
\end{tabular}
\end{table}

\begin{table}[htb]
\centering\small
\caption{What \sfthree{} inherits from and changes relative to
SceneFactory~\citep{scenefactory}. Entries for SceneFactory follow its paper.}
\label{tab:predecessor}
\begin{tabular}{@{}>{\raggedright\arraybackslash}p{0.19\linewidth}>{\raggedright\arraybackslash}p{0.36\linewidth}>{\raggedright\arraybackslash}p{0.37\linewidth}@{}}
\toprule
Aspect & SceneFactory & \sfthree{} \\
\midrule
Batching & Isaac Lab, many worlds per GPU & Inherited; Newton XPBD backend \\
Vehicle & Articulated PhysX body, no explicit tire model & Rigid chassis, raycast suspension, per-wheel tire forces (Sec.~\ref{sec:vehicle}) \\
Friction & One coefficient per world, from weather & Spatial patches evaluated at each wheel contact \\
Road geometry & Waymo topologies, flattened to $z=0$ & Per-world heightfields; procedural road in this study \\
Policy interface & Goal-reaching RL with weather context & Canonical routes shared by classical and learned controllers \\
Comparison & Friction-aware vs.\ friction-blind policy & Matched reference/changed worlds with recorded checks \\
\bottomrule
\end{tabular}
\end{table}

\section{Vehicle Model and Runtime}
\label{app:vehicle}
\paragraph{Simulation pipeline.} Newton integrates each chassis as a free rigid
body with XPBD (8 iterations, one substep per 240~Hz step) and resolves rigid
contacts between chassis and between chassis and terrain. A custom Warp kernel,
launched at every 240~Hz step, computes all other vehicle forces from the current
chassis state and applies their sum as an external wrench. Actions are updated at
60~Hz and held for four physics steps.

\paragraph{Suspension and contact.} Each corner has a massless raycast leg
anchored at $(\pm L/2,\pm t/2,-0.12)$~m in the body frame. A ray of length
$l_0+R$ along the chassis down axis queries the world's own road mesh. For
compression $\delta=l_0-(d-R)\ge0$ at hit distance $d$, the load is
$F_z=\max(0,k\delta-c\,v_n)+4k\max(0,\delta-0.8\,l_0)$, where $v_n$ is the contact
velocity along the surface normal; there is no contact when the ray misses or the
normal is steeper than $\cos^{-1}0.15$. The contact frame uses the surface normal
and the wheel heading projected onto the surface.

\paragraph{Tire and wheel.} With contact-frame velocities $(v_x,v_y)$, wheel spin
$\omega$, inertia $I_w$, drive torque $T_d$ and brake torque $T_b$ over step
$\Delta t$, the kernel uses brake damping $c_b=T_b/\max(|\omega|,0.5)$,
$D=I_w+\Delta t\,c_b$ and $k_x=70{,}000/\max(|v_x|,2)$, and computes
\begin{align*}
 F_x&=k_x\,\frac{R\,(I_w\omega+\Delta t\,T_d)/D-v_x}{1+k_xR^2\Delta t/D},&
 F_y&=-55{,}000\,\operatorname{atan2}\!\left(v_y,\max(|v_x|,2)\right),
\end{align*}
scales $(F_x,F_y)$ onto the circle of radius $\mu F_z$ with
$\mu=\mu_{\mathrm{vehicle}}\muenv(c)$ when it is exceeded, and updates
$\omega\leftarrow(I_w\omega+\Delta t(T_d-RF_x))/D$. Rolling resistance
$-0.015F_z\tanh(v_x/0.5)$ is then added longitudinally and the combined force is
clipped to the same circle. The longitudinal law is linear in slip velocity with
a speed-regularized stiffness, solved jointly with the wheel update; the lateral
law is a constant cornering stiffness applied to an arctangent slip angle. These
are standard small-slip approximations~\citep{pacejka2012} with hard
friction-circle saturation; the model has no force peak followed by a sliding
decline, no relaxation length and no load sensitivity other than the linear
bound. Aerodynamic drag $\tfrac12\rho C_dA\|v\|v$ acts on the chassis.

\paragraph{Actuation and command limits.} A requested acceleration $a$ maps to
throttle $\mathrm{clip}(a/a_{\mathrm{thr}},0,1)$ with
$a_{\mathrm{thr}}=T_{\max}/(Rm)=4.91$~m/s$^2$, and to brake
$\mathrm{clip}(-a/8,0,1)$. Drive torque is
$\min(T_{\max},P_{\max}/\max(\bar\omega,1))\cdot\mathrm{throttle}$, split equally
over the two rear wheels, where $\bar\omega$ is their mean spin; brake torque is
$\mathrm{brake}\cdot T_b$ on each wheel. The mapping is open loop, so realized and
requested acceleration differ. Normalized steering $s\in[-1,1]$ sets the target
$\delta^*=s\,\delta_{\max}/(1+(v/22)^2)$, approached at no more than 1.3~rad/s, with
Ackermann geometry for the two front wheels. Software limits act before the
dynamics: throttle, brake and steering are clipped to their ranges; the learned
policies command $a\in[-4,4]$~m/s$^2$ and full steering range; IDM-17 commands
$a\in[-6,a_{\max}]$ with $a_{\max}$ from the driver preset. Force saturation at the
tires is a separate, physical limit. Table~\ref{tab:vehicle} lists the parameters
of the single vehicle model used for every benchmark vehicle.

\begin{table}[htb]
\centering\small
\caption{Parameters of the benchmark vehicle model.}
\label{tab:vehicle}
\begin{tabular}{@{}lr@{\hskip 2em}lr@{}}
\toprule
Mass $m$ & 1400 kg & Spring rate $k$ & 34{,}000 N/m \\
Length $\times$ width & $4.25\times1.78$ m & Damping $c$ & 3{,}400 N\,s/m \\
Wheelbase $L$ / track $t$ & 2.62 / 1.54 m & Wheel inertia $I_w$ & 1.3 kg\,m$^2$ \\
Wheel radius $R$ & 0.32 m & Drive torque $T_{\max}$ (total) & 2{,}200 N\,m \\
Suspension rest length $l_0$ & 0.38 m & Power $P_{\max}$ & 135 kW \\
Grip factor $\mu_{\mathrm{vehicle}}$ & 1.05 & Brake torque $T_b$ (per wheel) & 1{,}800 N\,m \\
Steering limit $\delta_{\max}$ & 0.55 rad & Drag area $C_dA$ & 0.65 m$^2$ \\
Drive layout & rear-wheel & Air density $\rho$ & 1.225 kg/m$^3$ \\
\bottomrule
\end{tabular}
\end{table}

\paragraph{Provenance and scope.} The kernel is a GPU port of the platform's
single-vehicle reference implementation and is regression-tested against it. The
authored surface coefficient multiplies the vehicle grip factor, so it is not an
independently measured tire--road coefficient. Tire temperature, anti-lock
braking, detailed material response and deformable soil or snow are not modeled.

\paragraph{Batching and terrain ownership.} All worlds share one world-indexed
model; collisions are resolved within a world and never between worlds. Worlds
that share a terrain variant share its geometry. Each distinct terrain has a
wheel-query BVH and a rigid-contact BVH whose ownership must agree. Terrain is
fixed for an environment's lifetime; a partial reset restores the dynamic state
of the affected world and keeps its terrain. This does not restore every hidden
state needed to branch a warmed-up, continuing traffic experiment. The same
physical model also runs in an interactive Isaac Sim application for inspection;
the headless backend used for all experiments launches no rendering.

The frozen final-benchmark environment uses Newton 1.5.0, Warp 1.17.0,
PyTorch 2.10 with CUDA 12.8, Python 3.12 and Isaac Lab upstream commit
\texttt{28a37ce} with a compatibility overlay. The freeze manifest records
the benchmark code hash, checkpoint hashes, per-condition terrain hashes and
the canonical-route digest. Framework defaults and API changes can affect
physical behavior, so the release must preserve this tested configuration
rather than specify ``latest.''
\paragraph{Compute.} All benchmark, scaling and batch-sensitivity runs used a single
NVIDIA RTX PRO 6000 Blackwell Max-Q Workstation GPU (\ComputeGPUMem~GB, driver
580.173.02) in one workstation with an Intel Xeon w9-3575X CPU (\ComputeThreads{}
threads), \ComputeRAM~GB of memory and Ubuntu 24.04. Cells ran sequentially. The
wall time of all 252 Tier-2 cells, including 60~Hz logging of the rich subset,
evaluation and writing, was \ComputeTierTwoHours~h; the Tier-1 cells took
\ComputePDMHours~h for PDM-Closed-17, whose planner used 48 CPU workers, and
\ComputeIDMHours~h for IDM-17. Policy training is not included.

\section{Physical Verification}
\label{app:validation}
The checks below are implementation verification in the sense
of~\citet{sargent2013}: they establish that each authored intervention reaches
the implemented dynamics. They are internal-consistency checks rather than
findings or validation against real vehicles; most follow from the vehicle
model itself and would hold with no controller or traffic present. The closed-loop question of whether these interventions change
multi-agent execution and controller sensitivity is addressed in
Section~\ref{sec:results}; the quantities below serve there as mechanistic
context and interpretation boundaries.

Validation follows three levels. First, inspect the authored condition to
establish what changed. Second, query the loaded geometry and friction tables.
Third, run a vehicle probe that bypasses adaptive controller behavior, so a
controller cannot mask a missing physical intervention. Checks of learned
observations additionally compare preview values against the simulator's
physical tables and terrain geometry.

\begin{table}[htb]
\centering\small
\caption{Controller-free vehicle probes of all 21 conditions on the frozen
final-benchmark code. Wheel-contact $\mu$ is sampled under the four tire
contacts every step; slope is measured from the vehicle's own height profile.
Brake: mean deceleration under a full brake command from 16 m/s inside the
changed region, averaged while the front axle remains on it. Coast: zero command
from 12 m/s through the constant-grade section; the reference coast value
($-0.18$ m/s$^2$) reflects rolling and drivetrain resistance. Stop: distance from 16 m/s; ``exit'' gives the
speed at which the vehicle leaves the 140 m region without stopping. These are
internal consistency checks, not real-vehicle calibration.}
\label{tab:probes}
\begin{tabular}{@{}lrrrrr@{}}
\toprule
Condition & Wheel $\mu$ & Slope (\%) & Brake (m/s$^2$) & Stop (m) & Coast (m/s$^2$) \\
\midrule
Reference & 1.00 & 0.00 & 10.02 & 12.5 & $-0.18$ \\
\addlinespace
$\muenv=0.70$ & 0.70 & 0.00 & 7.23 & 17.3 & --- \\
$\muenv=0.50$ & 0.50 & 0.00 & 5.17 & 24.2 & --- \\
$\muenv=0.35$ & 0.35 & 0.00 & 3.63 & 34.5 & --- \\
$\muenv=0.25$ & 0.25 & 0.00 & 2.60 & 48.1 & --- \\
$\muenv=0.18$ & 0.18 & 0.00 & 1.88 & 66.5 & --- \\
$\muenv=0.12$ & 0.12 & 0.00 & 1.26 & 98.9 & --- \\
$\muenv=0.08$ & 0.08 & 0.00 & 0.86 & exit 7.1 m/s & --- \\
$\muenv=0.05$ & 0.05 & 0.00 & 0.57 & exit 11.0 m/s & --- \\
\addlinespace
Grade $+1.5\%$ & 1.00 & 1.50 & 10.14 & 12.2 & $-0.31$ \\
Grade $+3.0\%$ & 1.00 & 3.00 & 10.29 & 11.9 & $-0.45$ \\
Grade $+4.0\%$ & 1.00 & 4.00 & 10.39 & 11.7 & $-0.54$ \\
Grade $-1.5\%$ & 1.00 & $-1.50$ & 9.87 & 12.7 & $-0.04$ \\
Grade $-3.0\%$ & 1.00 & $-3.00$ & 9.72 & 13.0 & $0.10$ \\
Grade $-4.5\%$ & 1.00 & $-4.50$ & 9.57 & 13.3 & $0.24$ \\
Grade $-6.0\%$ & 1.00 & $-6.00$ & 9.42 & 13.6 & $0.38$ \\
Grade $-8.0\%$ & 1.00 & $-8.00$ & 9.24 & 14.0 & $0.56$ \\
\addlinespace
$-4.5\%$, $\muenv=0.35$ & 0.35 & $-4.50$ & 3.19 & 40.2 & $0.24$ \\
$-4.5\%$, $\muenv=0.18$ & 0.18 & $-4.50$ & 1.44 & 88.7 & $0.24$ \\
$+3.0\%$, $\muenv=0.35$ & 0.35 & 3.00 & 3.92 & 31.4 & $-0.45$ \\
$+3.0\%$, $\muenv=0.18$ & 0.18 & 3.00 & 2.17 & 56.7 & $-0.45$ \\
\bottomrule
\end{tabular}
\end{table}

In all 21 conditions the sampled wheel-contact coefficient equals the authored
value, and the measured slope matches the authored grade to within $0.05$
percentage points. Achieved braking deceleration corresponds to an effective
coefficient of 1.02--1.07 times the authored value for $\muenv\geq0.12$,
rising to 1.10 and 1.15 at 0.08 and 0.05, consistent with a roughly fixed
additional resistance. The coast acceleration shifts by approximately
$0.94\,g\cdot\mathrm{grade}$ with the correct sign, and braking is
direction-aware. Composed conditions show both interventions at once:
a $-4.5\%$ descent lengthens the $\muenv=0.18$ stop by about a third, whereas
a $+3.0\%$ climb offsets part of the friction loss.

Two properties of this model matter for interpreting controller results.
First, because Equation~\ref{eq:friction} scales forces only when they exceed
$\mu F_z$, the friction coefficient does not enter the computation at all while
demand stays below the bound. A 0.6~m lateral route offset tracked with the
Stanley law at 17~m/s therefore produces bit-identical trajectories for every
$\muenv\geq0.12$, since its lateral demand of about 0.95~m/s$^2$ stays below
$\mu g$, while at $\muenv=0.05$ the same maneuver leaves the lane. This is a
property of this implementation and probe, not of tire mechanics in general.
Second, the conditions differ widely in physical magnitude. The steepest grade,
$-8\%$, changes full-brake deceleration by about 8\%, whereas $\muenv=0.25$
removes about 74\%. For a vehicle that brakes fully on entering the patch at
16~m/s, the stopping distance is 98.9~m at $\muenv=0.12$, within the 140~m
patch, whereas at 0.08 and 0.05 it leaves the patch at 7.1 and 11.0~m/s; the
probe vehicle occupies the closing lane, on which the patch lies. Under a
$+3$~m/s$^2$ command at $\muenv=0.05$ the measured net forward acceleration is
0.11~m/s$^2$. These are limits for the probed entry states: a controller that
slows before the patch, or completes its merge onto the dry open lane, faces
different ones, so the probes do not bound task performance.
The four-condition probe configuration used during development ($-6$~m/s$^2$
command from 16~m/s just inside the region) reproduces its recorded stopping
distances to the centimeter on the frozen code.

Separate training-environment checks compare route-aligned preview against
the terrain sampled by physics. The recorded check reports exact agreement
for friction at 10,902 sampled preview points and maximum grade discrepancy
of $2.0\times10^{-5}$ under its stated finite-difference convention. Encounter
checks also identify patch-edge discrepancies, so point-query agreement does
not mean that every wheel simultaneously experiences the previewed surface.
The zero-distance query is at the route projection; wheel contacts can straddle
a boundary or depart from the route.

Terrain-collider validation covers 1, 2, 4, 16 and 64 simultaneous worlds, up
to 64 unique 43,200-triangle terrains. It checks collider geometry, wheel-query
geometry and the world's assigned variant, together with physical chassis
contact and partial resets. These checks
exclude two failure modes that a terrain label or rendering would not reveal:
per-world wheel queries paired with a shared chassis collider, and automatic
convexification of the road mesh.

\section{Controller Interfaces and Training}
\label{app:controllers}
\paragraph{Routes.} The curvature budget is
$\kappa_{\max}=\min\!\big(0.5\tan(\delta_{\max}/(1+(12/22)^2))/L,\;2/12^2\big)=0.0139$~m$^{-1}$,
i.e.\ half the steering authority available at a 12~m/s design speed, capped at
2~m/s$^2$ of lateral acceleration. Controllers project the vehicle onto the
route at its nearest 0.5~m sample within a local window around the current arc
length. Controllers decide on the route geometry; the closure enters only
through the route and the outcome rules.

\paragraph{Classical controllers.} IDM-17 applies the IDM
law~\citep{idm} to a route-relative leader (the nearest vehicle ahead on the
vehicle's own route within 2~m laterally), with each vehicle's preset headway,
minimum gap and acceleration limits and a desired speed of 17~m/s, and follows
the route laterally with a Stanley law~\citep{stanley}. It does not consult a
lane-change model. PDM-Closed-17 runs the published
implementation~\citep{pdm} (upstream commit \texttt{b51d5d0}) with the canonical
route as its centerline, lateral offsets of $\pm1$~m, speed fractions
$\{0.2,\dots,1.0\}$ of a 17~m/s road limit, its own proposal scoring and
kinematic forecast, and its LQR tracker; the adapter converts states to its
rear-axle convention and builds the drivable area from the road minus the closed
lane. The two classical controllers differ in lateral tracking and decision rate,
so their comparison is not an ablation of proposal search. PDM-Closed-17 and the
learned policies decide at 10~Hz, IDM-17 at 60~Hz; all feed the 60~Hz actuation
bridge. The 17~m/s targets reduce, but do not remove, differences in realized
speed.

\begin{table}[htb]
\centering\small
\caption{Controller regimes. Each regime drives all 12 vehicles of a world, with
the same canonical routes, traffic state and vehicle model. Preview is
privileged, noise-free physical state sampled along the route.}
\label{tab:controllers}
\begin{tabular}{@{}>{\raggedright\arraybackslash}p{0.20\linewidth}>{\raggedright\arraybackslash}p{0.41\linewidth}>{\raggedright\arraybackslash}p{0.16\linewidth}>{\raggedright\arraybackslash}p{0.13\linewidth}@{}}
\toprule
Controller & Decision mechanism & Terrain training & Preview \\
\midrule
IDM-17 & IDM longitudinal law + Stanley route tracking & --- & No \\
PDM-Closed-17 & Published PDM proposals/scoring + LQR tracking & --- & No \\
\Lzero & Route-conditioned policy, nominal roads & No & No \\
\Lzerot & \Lzero{} continued on terrain-diverse roads & Yes & No \\
\Lone & As \Lzerot{}, with physical preview inputs & Yes & Yes \\
\bottomrule
\end{tabular}
\end{table}

\paragraph{Learned observation.} Table~\ref{tab:obs} lists the 112 channels
shared by all learned policies and the 12 channels only \Lone{} receives. All
channels are divided by the fixed scales shown and clipped to $\pm10$; there is
no learned observation normalizer. The 24-channel planner-semantic block exists
in the input layout of the parent family but is held at zero in training and
evaluation. No channel carries a condition label, terrain identifier or future
vehicle state.

\begin{table}[htb]
\centering\small
\caption{Learned-policy observation, ego frame, divided by the listed scale.}
\label{tab:obs}
\begin{tabular}{@{}>{\raggedright\arraybackslash}p{0.18\linewidth}r>{\raggedright\arraybackslash}p{0.66\linewidth}@{}}
\toprule
Block & Width & Content (scale) \\
\midrule
Ego & 9 & speed (12), lateral speed (5), yaw rate (1), steering angle (0.55), previous acceleration and steering actions (1), cross-track error (2), heading error (1), route curvature (0.02) \\
Route preview & 12 & route points 2, 5, 10, 15, 20, 30~m ahead, forward and left (30) \\
Acceleration & 1 & speed change over the last decision divided by 0.1~s (4) \\
Route leader & 6 & presence, gap (50), relative speed (12), lateral offset (2), sine and cosine of relative heading \\
Neighbors & 60 & six nearest vehicles within 40~m, each: valid, $x$ (30), $y$ (10), $v_x$ (12), $v_y$ (5), sine and cosine of relative heading, distance (30), route-arc offset (30), lateral offset (5) \\
Semantics & 24 & held at zero \\
\midrule
\Lone{} friction & 6 & $(1-\muenv)/0.5$ at route points 0, 5, 10, 20, 30, 40~m ahead \\
\Lone{} grade & 6 & $(dz/ds)/0.05$ at the same points, central difference over $\pm2.5$~m along the route heading \\
\bottomrule
\end{tabular}
\end{table}

\paragraph{Policy, action and reward.} Actor and critic are separate
256--128--64 ELU networks that receive the same observation; the critic has no
privileged input. The actor outputs a Beta distribution on
$(-1,1)^2$~\citep{chou2017}; an action $(a_0,a_1)$ is decoded as acceleration
$4a_0$~m/s$^2$ and normalized steering $a_1$. Training samples actions; every
evaluation uses the distribution mean. One action is held for six 60~Hz control
steps. The per-decision reward of a controlled vehicle is
\begin{equation*}
 r=\Delta s^{+}-0.1\left(2e^2+\psi^2+10\,\mathbb{1}_{\mathrm{off}}\right)
 -25\,\mathbb{1}_{\mathrm{fault}}
 -\Big(5\,n_{\mathrm{onset}}+60\sum_k\rho_k+t_{\mathrm{contact}}\Big)
 -4.25\max_{j}\mathrm{risk}_j,
\end{equation*}
where $\Delta s^{+}$ is new route progress beyond the vehicle's previous frontier
(zero while touching another vehicle or more than 1.5~m from the route), $e$ and
$\psi$ are cross-track and heading errors, $\mathbb{1}_{\mathrm{off}}$ marks the
body extent more than 2.5~m from the route, $n_{\mathrm{onset}}$ counts new contact
onsets, $\rho_k\in[0,1]$ is the vehicle's directional responsibility for onset
$k$, $t_{\mathrm{contact}}$ is the time in contact in seconds, and
$\mathrm{risk}_j=\exp(-(g_j^*/0.6)^2)\exp(-t_j^*/1.5)$ uses the minimum predicted
gap $g_j^*$ between vehicle rectangles and its time $t_j^*$ under a
constant-velocity extrapolation of neighbor $j$ over 3~s, computed from current
states only. A vehicle's episode ends on success (route end reached with
maximum error below 1.5~m, heading error below 0.3~rad and no fault), on failure
(more than 6~m from the route or a fault), or at the 35~s time limit, where the
critic bootstraps. Collisions do not terminate episodes.

\paragraph{Optimization and budget.} PPO~\citep{ppo} uses 64 decisions per
environment per update, 4 mini-batches, 5 epochs, clip 0.2, entropy coefficient
0.005, value-loss coefficient 1.0 with clipping, gradient-norm limit 1.0, and
discount and GAE parameters $0.997$ and $0.97$ per 60~Hz step, raised to the sixth
power per decision. Continuation uses a fixed learning rate of $3\times10^{-4}$
in three stages whose final checkpoints advance without selection on evaluation
data: 100 updates with one vehicle in each of 1{,}024 worlds on single-route
scenes, 100 updates on traffic scenes with 256 worlds of six vehicles, and 300
updates on the parent's full scene mixture with 256 worlds of six vehicles. This
is at most $4.6\times10^{7}$ vehicle decisions per seed (4.6 million vehicle-seconds),
an upper bound because inactive rows contribute no decisions. Each continuation
seed starts from the parent checkpoint with the same seed and a fresh optimizer
in the first stage. Because there is no learned normalizer, zero-initializing
the 12 added input columns ($12\times256=3{,}072$ first-layer actor weights, and
the same for the critic) makes \Lone{} compute exactly its parent's function at
the start; the architectures thus differ only in these input weights. \Lzerot{}
and \Lone{} share stages, scenes, terrain sampling, seeds and reward. The
parents' own training budget is recorded in the parent training reports and is
identical across all arms.

\paragraph{Training and test support.} Table~\ref{tab:support} compares the
continuation training distribution with the benchmark. Terrain is sampled per
episode from a stream independent of the scene generator; 60\% of patches are
placed near the scene's maneuver point.

\begin{table}[htb]
\centering\small
\caption{Training support of the terrain-diverse continuation versus the final benchmark.}
\label{tab:support}
\begin{tabular}{@{}>{\raggedright\arraybackslash}p{0.2\linewidth}>{\raggedright\arraybackslash}p{0.4\linewidth}>{\raggedright\arraybackslash}p{0.32\linewidth}@{}}
\toprule
Variable & Continuation training & Final benchmark \\
\midrule
Friction mode & none 40\%, road band 40\%, lane band 10\%, whole road 10\% & closing-lane patch \\
Friction value & bands $\muenv\sim U(0.2,0.7)$; whole road $U(0.35,0.8)$ & 0.70 to 0.05 \\
Patch length & 25--70~m (25\% add a second band) & 140~m \\
Grade & half flat, half rolling fields: three plane waves, wavelength 90--220~m, maximum grade 4--9\% & constant $+4\%$ to $-8\%$ with 20~m transitions \\
Road and task & procedural single routes, following, platoons, lane shifts, merges, crossings & straight two-lane closure \\
Vehicles per world & 1 (first stage) or 6 & 12 \\
Initial states & scene generator seeds disjoint from evaluation & held-out closure seeds \\
\bottomrule
\end{tabular}
\end{table}

The benchmark therefore combines three kinds of novelty: unseen road geometry and
task, a larger population, and, below $\muenv=0.2$, friction values outside the
training range. The preview is privileged and exact, without perception error,
latency or occlusion.

\section{Conditions, Outcomes and Statistics}
\label{app:conditions}
\paragraph{Conditions and geometry.} The frozen registry
(\texttt{ucx-conditions-1}) has 21 conditions: the reference road
($\muenv=1.0$); eight friction levels (0.70, 0.50, 0.35, 0.25, 0.18, 0.12, 0.08,
0.05); eight grades ($+1.5$, $+3.0$, $+4.0$, $-1.5$, $-3.0$, $-4.5$, $-6.0$,
$-8.0\%$, positive uphill in the direction of travel); and four compositions of
grades $-4.5\%$ and $+3.0\%$ with friction 0.35 and 0.18. Both tiers use it
unchanged. Friction patches cover the closing lane for $x\in[40,180]$~m; grade
regions start at $x=40$~m, span the road width, and have 20~m vertical
transitions around a constant-grade section. All conditions share flat reference
ground upstream of $x=40$~m. The road mesh spans 2{,}560~m with 5~m station
spacing; the closure starts at $x=150$~m, the closing-lane route merges between
50 and 110~m, and the counting line lies at $x=170$~m. The horizon is 66~s (3{,}960
control steps).

\paragraph{Worlds and tiers.} Tier~2 uses scenario seeds 100000--101023; each
checkpoint--condition cell is one colocated batch of 1{,}024 worlds. Tier~1 uses the
first 32 of these worlds: the classical controllers run on them as 32-world batches,
and the learned controllers' Tier-1 outcomes are the first 32 worlds of their
Tier-2 cells, an exact subset rather than a rerun (initial-state digests are
identical). The first 64 worlds of every Tier-2 cell were also recorded at 60~Hz
from the same rollout (the rich subset). Reruns of the same batch reproduce
outcomes almost exactly, but a change of batch size perturbs individual worlds
(Appendix~\ref{app:final}); this checks repeatability on one machine, not bitwise
determinism across hardware or software versions.

\paragraph{Outcome rules.} All rules read the recorded trajectory at 60~Hz and no
controller state. For each vehicle:
\begin{itemize}
\setlength{\itemsep}{1pt}
\item \emph{Closure incursion}: the body center lies in the closed lane at
$x\geq150$~m for at least one step. A merging vehicle whose tail sweeps over the
lane line while its center is upstream does not qualify.
\item \emph{Collision}: the vehicle's footprint (4.25$\times$1.78~m, grown by
0.05~m) intersects another vehicle's, by a separating-axis test.
\item \emph{Off-road}: the body center leaves the paved road
($y\notin[-31.5,-24.5]$~m).
\item \emph{Fault}: the vehicle overturns (body up-axis $z$-component below 0.2),
leaves the map bounds, falls below the terrain, or has a non-finite state; a
world reset (below) is a fault for every vehicle in that world from that step on.
\item \emph{Validity}: true until the first step with any of the four events above,
counted on the step it occurs, and never restored. The first event determines the
reported failure cause, with ties broken in the order fault, collision, off-road,
incursion.
\item \emph{Valid passage}: the front bumper crosses $x=170$~m within the horizon
while the vehicle is valid. Vehicles continue driving, interacting and being
scored for events after they cross; an event after the crossing does not revoke
valid passage.
\item \emph{Merge completion}: for a closing-lane vehicle, its footprint lies
entirely in the open lane from some step until the horizon.
\item \emph{Strict task success}: valid passage, merge completion if one is owed,
and no event of any kind over the whole horizon.
\end{itemize}
Per-world rates average these indicators over all 12 vehicles; the denominator is
always 12 and no world or vehicle is discarded.

\paragraph{Resets, censoring and map edge.} When any vehicle faults, its whole
world terminates and restarts. Nothing logged after a world's first reset is
scored. A vehicle that is still valid and has not crossed at that step is
\emph{censored}; it counts as a non-pass in the fixed denominator, and censoring
rates are reported per controller and condition. The first reset can therefore
truncate the outcomes of vehicles other than the one that caused it, and censoring
can depend on controller and condition. The finite mesh adds one controller-independent
geometric rule: a map-adjusted success measure does not count a vehicle's own
termination as a failure if it occurred within 2~m of the mesh edge, while still
requiring its crossing before the first reset. This adjustment exists because a
controller that ignores the scenario's desired speeds can finish the task and
reach the downstream mesh edge before the horizon. In the final
benchmark, map-edge resets account for 0.06\% of censored vehicles, so the
adjustment changes no reported result. The released evaluator's aggregate
collision, off-road and censoring rates also read the log after a reset; every
number in this paper is recomputed on the first episode
(Appendix~\ref{app:final}).

\paragraph{Sensitivity and paired decomposition.} For controller $\pi$ and
condition $p$, the sensitivity is the mean over worlds of the per-world
difference in valid-passage rate between $p$ and the reference, computed on the
same worlds. It is reported with the reference and changed-condition rates.
Any paired descriptive decomposition (passes preserved, lost, gained, and
persistent non-passes) must use the full fixed set of worlds and vehicles, never a
controller-specific subset of nominally successful worlds.

\paragraph{Statistics.} Worlds are the resampling unit. For one frozen checkpoint,
an estimate is the mean of per-world values, and its interval is a percentile
bootstrap over worlds (2{,}000 replicates, 95\%) applied to per-world paired
differences, so the matched sides are never resampled independently. This interval
is conditional on the checkpoint and treats worlds as draws from the scenario
generator. Training variability is described by the four per-seed estimates:
mean, standard deviation and a $t$-interval with three degrees of freedom, which
assumes approximately normal seed-level estimates. Family contrasts pair \Lzerot{}
and \Lone{} seed $i$ with their common \Lzero{} parent $i$ and compute per-world
differences before averaging. Because the same worlds are shared across seeds,
arms and conditions, a nested bootstrap resamples seeds and then one common set of
world indices for all seeds in a replicate; it is reported beside, not instead of,
the seed-level summary, and it does not create additional independent training
runs. Intervals on severity curves are pointwise, not simultaneous. Exposure-normalized
safety rates are ratios of sums over worlds, recomputed in every replicate.
Bootstrap seeds are fixed and recorded with the analysis outputs.

\section{Studies and Preliminary Evidence}
\label{app:preliminary}
Table~\ref{tab:studies} separates the studies reported in this paper. Only the two
final tiers use the benchmark task; the development study uses held-out scenes of
the training scene families and different outcome definitions.

\begin{table}[htb]
\centering\small
\caption{Studies, their unit of analysis and their role.}
\label{tab:studies}
\begin{tabular}{@{}>{\raggedright\arraybackslash}p{0.15\linewidth}>{\raggedright\arraybackslash}p{0.33\linewidth}>{\raggedright\arraybackslash}p{0.14\linewidth}>{\raggedright\arraybackslash}p{0.12\linewidth}>{\raggedright\arraybackslash}p{0.14\linewidth}@{}}
\toprule
Study & Task and controllers & Sample & Horizon & Role \\
\midrule
Probes & One vehicle, fixed commands, 21 conditions & one run per condition & to stop or exit & verification (App.~\ref{app:validation}) \\
Development & Held-out training-family scenes; three learned families, 4 seeds, plus nominalized preview & 256 scenes $\times$ 6 vehicles per cell & 35~s & preliminary evidence \\
Final Tier 1 & Closure task; all five regimes; 21 conditions & 32 worlds $\times$ 12 vehicles & 66~s & cross-controller comparison \\
Final Tier 2 & Closure task; three learned families, 4 seeds; 21 conditions & 1{,}024 worlds $\times$ 12 vehicles & 66~s & learned-controller statistics \\
\bottomrule
\end{tabular}
\end{table}

Table~\ref{tab:development} summarizes the development study's flat-ground
evaluation as means and standard deviations over four trained seeds. Route
success requires reaching the route end with maximum error below 1.5~m, heading
error below 0.3~rad and no fault; failure denotes leaving the route by more than
6~m or a physics fault. The two are not complements, and neither is the work-zone
valid passage. The conditions are an all-lane friction band, a lane-local patch
and a whole-road friction change, all within the continuation training range. The
nominalized-preview rows keep the trained \Lone{} weights and physical world but
replace the physical observations with nominal values. Standard deviations over
seeds are not confidence intervals.
\begingroup\small
\begin{longtable}{@{}llrrr@{}}
\caption{Preliminary controller-development evaluation: mean $\pm$ standard deviation
over four training seeds. Success and failure are route-task metrics defined in
Appendix~\ref{app:preliminary}. Speed is included to expose operating-point differences.
These are not final two-tier benchmark results.}\label{tab:development}\\
\toprule Condition & Controller & Success & Failure & Speed (m/s) \\
\midrule\endfirsthead
\multicolumn{5}{l}{\textit{Table \thetable{} continued: preliminary development evaluation}}\\
\toprule Condition & Controller & Success & Failure & Speed (m/s) \\
\midrule\endhead
\bottomrule\endfoot
Nominal & Frozen L0 & $0.991 \pm 0.008$ & $0.002 \pm 0.003$ & $12.37 \pm 0.84$ \\
 & L0-T & $0.993 \pm 0.007$ & $0.001 \pm 0.002$ & $11.99 \pm 1.40$ \\
 & L1 & $0.994 \pm 0.003$ & $0.002 \pm 0.001$ & $11.81 \pm 1.63$ \\
 & L1 (nominal preview) & $0.994 \pm 0.003$ & $0.002 \pm 0.001$ & $11.81 \pm 1.63$ \\
\midrule
Road band $\mu=0.50$ & Frozen L0 & $0.977 \pm 0.014$ & $0.009 \pm 0.006$ & $12.16 \pm 0.79$ \\
 & L0-T & $0.987 \pm 0.011$ & $0.003 \pm 0.003$ & $11.80 \pm 1.33$ \\
 & L1 & $0.992 \pm 0.006$ & $0.002 \pm 0.001$ & $11.49 \pm 1.30$ \\
 & L1 (nominal preview) & $0.988 \pm 0.010$ & $0.003 \pm 0.002$ & $11.63 \pm 1.52$ \\
\midrule
Road band $\mu=0.35$ & Frozen L0 & $0.876 \pm 0.087$ & $0.046 \pm 0.027$ & $11.32 \pm 0.43$ \\
 & L0-T & $0.949 \pm 0.043$ & $0.011 \pm 0.009$ & $11.21 \pm 1.07$ \\
 & L1 & $0.956 \pm 0.043$ & $0.010 \pm 0.008$ & $10.94 \pm 1.00$ \\
 & L1 (nominal preview) & $0.938 \pm 0.066$ & $0.017 \pm 0.014$ & $11.06 \pm 1.17$ \\
\midrule
Road band $\mu=0.25$ & Frozen L0 & $0.770 \pm 0.095$ & $0.085 \pm 0.025$ & $10.50 \pm 0.26$ \\
 & L0-T & $0.887 \pm 0.052$ & $0.025 \pm 0.011$ & $10.63 \pm 0.90$ \\
 & L1 & $0.897 \pm 0.055$ & $0.024 \pm 0.011$ & $10.35 \pm 0.76$ \\
 & L1 (nominal preview) & $0.880 \pm 0.078$ & $0.029 \pm 0.017$ & $10.46 \pm 0.92$ \\
\midrule
Lane patch $\mu=0.30$ & Frozen L0 & $0.908 \pm 0.055$ & $0.030 \pm 0.015$ & $11.54 \pm 0.52$ \\
 & L0-T & $0.973 \pm 0.021$ & $0.008 \pm 0.006$ & $11.38 \pm 1.15$ \\
 & L1 & $0.969 \pm 0.023$ & $0.006 \pm 0.002$ & $11.16 \pm 1.16$ \\
 & L1 (nominal preview) & $0.966 \pm 0.032$ & $0.006 \pm 0.006$ & $11.21 \pm 1.24$ \\
\midrule
Whole road $\mu=0.40$ & Frozen L0 & $0.768 \pm 0.192$ & $0.080 \pm 0.072$ & $10.62 \pm 0.39$ \\
 & L0-T & $0.962 \pm 0.029$ & $0.005 \pm 0.004$ & $10.85 \pm 0.78$ \\
 & L1 & $0.969 \pm 0.023$ & $0.005 \pm 0.003$ & $10.82 \pm 0.55$ \\
 & L1 (nominal preview) & $0.933 \pm 0.079$ & $0.009 \pm 0.009$ & $10.72 \pm 0.86$ \\
\end{longtable}
\endgroup

On a road band with $\muenv=0.25$, mean failure rates were
\DevBandFailureFrozen{} for \Lzero{}, \DevBandFailureTerrain{} for \Lzerot{} and
\DevBandFailurePreview{} for \Lone{} over four seeds. This friction lies inside the
continuation training range, so for the terrain-trained arms the shifted variable
is the held-out scene, not the friction amplitude. On rolling ground in the same study, grade alone changes completion little, and
rolling ground combined with reduced friction again degrades the nominal policy
more than the terrain-trained families. In a transfer of the same checkpoints to
development seeds of the closure task, some terrain-trained seeds showed higher
collision rates than their parents on the reference road; additional physical
training therefore need not preserve every nominal capability.

\section{Preview Analyses}
\label{app:preview}
All analyses use the development study's held-out scenes and fixed checkpoints;
no policy is retrained. \emph{Input interventions} change only what a fixed
\Lone{} policy observes: the nominalized preview replaces all 12 physical channels
with their nominal (zero) values; the current-only and ahead-only variants keep the
0~m or the 5--40~m channels; shifted previews report the terrain 20~m ahead of
(early) or behind (late) the true query points; a false preview reports a constant
$\muenv=0.3$. Paired differences use 256 scenes $\times$ 4 seeds with a bootstrap
over scenes pooled across seeds. Nominalizing the preview lowered success by
0.016--0.038 (resolved) on the $\muenv=0.35$ and $0.25$ bands and whole-road
$\muenv=0.4$, and changed nothing resolvable on the lane patch. The late shift
reproduced the nominalized outcome, the early shift performed about as well as the
true preview, and the constant false preview was about as safe as the true
one on these patch-bearing scenes. Because nominalized or false context disagrees
with the dynamics the vehicle experiences, these interventions measure the policy's
dependence on its inputs, not the value of correct information.

\emph{Behavioral analyses} align each policy vehicle that starts at least 45~m
before a patch (844 vehicles per band condition over four seeds) on route distance
to the patch entry and average with equal weight per vehicle. At the $\muenv=0.25$
band, \Lone{} accelerated more than its nominalized self 30~m before the patch
($+1.49$ versus $+1.31$~m/s$^2$) and less at the entry ($+0.67$ versus
$+1.06$~m/s$^2$), entering at the same speed as \Lzerot{} (13.9 versus 13.7~m/s).
Regressing commanded acceleration on grade under the vehicle and on the change in
grade over the next 20~m, on rolling ground without patches, gave positive
coefficients on the grade ahead for three of four \Lone{} seeds and negative ones
for every seed of the other arms and of nominalized \Lone{}. A sensitivity probe
that overwrites only the terrain channels on 4{,}000 recorded decision states per
seed found no anticipatory braking for friction reduced only ahead of the vehicle.

The \emph{training-regime comparison} of \Lone{} with \Lzerot{} is a different
question: it compares two policies trained identically except for the preview.
In the development study, success and failure did not differ resolvably on any
friction condition; the only resolved safety difference was small (at-fault
collision onsets lower by 0.005 per vehicle at whole-road $\muenv=0.4$), and
\Lone{} drove 0.13--0.33~m/s slower in most conditions. The final benchmark
result is reported in Section~\ref{sec:results}.

\section{Computational Scaling}
\label{app:scaling}
The study distinguishes $N$, the number of worlds; $A$, vehicles per world; and
$K$, unique terrain geometries. All runs use the production controller (\Lone{}
seed~0), $A=12$ live vehicles per world, development scenario seeds disjoint from
the benchmark, one NVIDIA RTX PRO 6000 Blackwell GPU and the frozen benchmark code,
with three repeats per point, each on a fresh environment. The shared-terrain
workload ($K=1$) places every world on the reference road and is run without
logging, with the full 60~Hz log, and with log plus safety diagnostics (the
production configuration). The heterogeneous workload places world $i$ on
condition $i \bmod 21$ ($K=21$) in the production configuration.

For $S=3{,}960$ timed control steps (66~s at 60~Hz, four 240~Hz physics substeps
each) and loop wall time $t$, world-step throughput is $NS/t$ and vehicle-step
throughput is $NAS/t$; all $A$ vehicles are live and stepped in every world. The
timed loop contains physics, observation, policy inference, actuation and, where
enabled, logging and safety diagnostics. It excludes environment and terrain
construction (0.5--3.7~s), a 2~s warm-up
rollout, reset and placement, device-to-host transfer, outcome evaluation and
rendering (none). Peak memory is the PyTorch allocator's peak. PDM-Closed-17 is not
part of this sweep: its planner runs on CPU workers and would measure the planner, not the
batched simulation.

\begin{table}[htb]
\centering\small
\caption{Throughput and memory against world count (means of three repeats;
repeat standard deviations are at most 2\% of each mean). Efficiency is throughput
relative to 64 worlds divided by the ideal ratio $N/64$.}
\label{tab:scaling}
\begin{tabular}{@{}rccccccc@{}}
\toprule
 & \multicolumn{3}{c}{Vehicle-steps/s (millions)} & 21 terrains & Loop (s) & Peak GPU & Efficiency \\
\cmidrule(lr){2-4}
Worlds & no log & log & log+safety & log+safety & log+safety & (GB) & log+safety \\
\midrule
64 & 0.15 & 0.13 & 0.11 & 0.10 & 27.0 & 0.56 & 1.00 \\
128 & 0.30 & 0.25 & 0.23 & 0.19 & 27.0 & 1.10 & 1.00 \\
256 & 0.59 & 0.51 & 0.45 & 0.39 & 26.9 & 2.18 & 1.00 \\
512 & 1.15 & 0.99 & 0.89 & 0.75 & 27.3 & 4.35 & 0.99 \\
1{,}024 & 2.24 & 1.92 & 1.71 & 1.40 & 28.4 & 8.67 & 0.95 \\
\bottomrule
\end{tabular}

\end{table}

\begin{figure}[htb]
\centering
\includegraphics[width=0.85\linewidth]{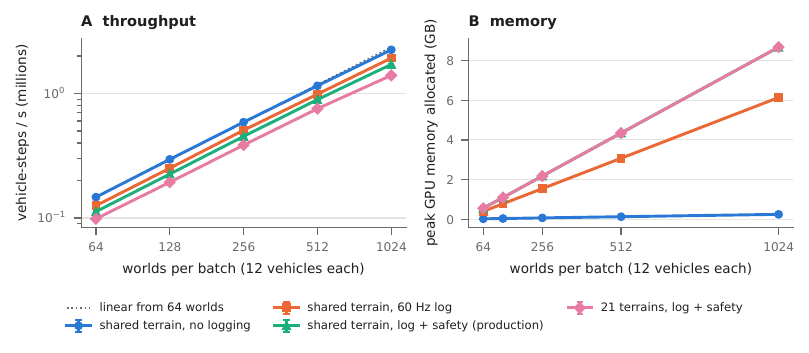}
\caption{Throughput (A) and peak GPU memory (B) against the number of worlds
on one GPU.}
\label{fig:scaling}
\end{figure}

Over 64--1{,}024 worlds the loop time is nearly constant (Table~\ref{tab:scaling}),
so throughput grows almost in proportion to $N$: in this range the loop is
dominated by per-step costs that do not grow with the batch. The measurements show headroom up to
1{,}024 worlds on this GPU; they do not locate the saturation point, and they do
not vary $A$.

\paragraph{Controller cost in production.} The production manifests time the same
loop for every cell (median over cells). A learned policy took
\LearnedLoopPerWorld~s of loop time per 66-s world in 1{,}024-world batches and
\LearnedLoopPerWorldSixtyFour~s at 64 worlds in the scaling study. IDM-17 took
\IDMLoopPerWorld~s per world in its 32-world batches, where the GPU loop is dominated by
fixed per-step costs. PDM-Closed-17 took \PDMLoopPerWorld~s per world
(\PDMVehSteps{} vehicle-steps/s) with 48 CPU planning workers, and the profiled steps
attribute almost all of this to its planner: about \PDMSlowdownSixtyFour{} times the
learned cost at a comparable batch size and \PDMSlowdownProd{} times the production
learned cost. This is the cost of the published planner in our adaptation, not of
the simulator, and it is why PDM-Closed-17 appears only in Tier~1.

\section{Production Audit and Supplementary Results}
\label{app:final}
\paragraph{Production audit.} The frozen production (metric schema
\texttt{safety-metrics-v2}; benchmark code hash \texttt{77d72a10\ldots}; 12
checkpoint, 21 terrain and 2 scenario digests fixed in the freeze manifest) was
checked cell by cell. All 252 Tier-2 cells, all 252 rich-subset records and all 42
Tier-1 cells are complete; every manifest matches the frozen code hash, schema,
terrain, checkpoint, seed list and scenario digest; no cell failed, was discarded or
was restarted; and in every cell the evaluator's valid passage equals the count of
passages before the first reset. The Tier-1 classical cells ran in two stages
separated by the Tier-2 run, as scheduled. An earlier safety schema, whose
quantities accumulated after failure, is archived separately and not used. No
equal-budget nominal continuation was run.

\paragraph{First-episode outcomes.} After a world reset the environment starts a
new episode and the log continues. Valid passage is unaffected, because a reset
invalidates every vehicle. The evaluator's aggregate collision, off-road and
censoring rates, however, read the whole log; for example, \Lzero{} seed~0 at
$\muenv=0.25$ has an aggregate collision rate of 0.33 but a first-episode collision
rate of 0.08. All outcome classes reported here (own collision, off-road,
incursion or termination as the first failure; censoring by another vehicle's
reset; valid but not yet crossed at the horizon) are recomputed from first-event
steps before the first reset.

\paragraph{Safety windows.} Within the episode, post-failure steps still inflate
interaction measures: at $\muenv=0.08$ the time below TTC 1.5~s per second of
driving is 1.7--2.3 times larger over the whole episode than over the valid-driving
window used here.

\paragraph{Batch-size sensitivity.} The same 64 worlds (\Lone{} seed~0,
reference and $\muenv=0.18$) were embedded in batches of 64, 128, 256, 512 and
1{,}024 worlds, with repeats at 64 and 1{,}024, and compared with the production
cell. The criterion, fixed before the comparison, required cross-batch differences
to stay within the same-batch rerun envelope; it is not met
(Table~\ref{tab:batch}). The mean shifts have both signs, and the
reference-to-$\muenv=0.18$ change on these worlds ranges from $-0.09$ to $-0.13$
across batch sizes.

\begin{table}[htb]
\centering\small
\caption{Batch-size sensitivity on 64 common worlds (768 vehicles).}
\label{tab:batch}
\begin{tabular}{@{}lcccc@{}}
\toprule
Condition & Same-batch reruns & \multicolumn{3}{c}{Across batch sizes 64--1{,}024} \\
\cmidrule(lr){2-2}\cmidrule(lr){3-5}
 & vehicle flips (max) & vehicle flips & worlds changed & $|\Delta$ mean VP$|$ (max) \\
\midrule
Reference & 0 of 768 & 11--28 of 768 & 6--14 of 64 & 0.010 \\
$\mu=0.18$ & 2 of 768 & 70--112 of 768 & 22--34 of 64 & 0.029 \\
\bottomrule
\end{tabular}

\end{table}

\begin{table}[htb]
\centering\scriptsize
\setlength{\tabcolsep}{2.5pt}
\caption{Tier-2 valid passage on the reference road and its change for selected
conditions, per checkpoint (1{,}024 worlds). Bold: the 95\% paired world-bootstrap
interval excludes zero. Friction columns give $\muenv$; grade columns the slope;
the last two are compositions with $\muenv=0.18$.}
\label{tab:checkpoints}
\resizebox{\linewidth}{!}{\begin{tabular}{@{}lcccccccccccccc@{}}
\toprule
Checkpoint & Ref. & ..70 & ..50 & ..35 & ..25 & ..18 & ..12 & ..08 & ..05 & $+4\%$ & $-4.5\%$ & $-8\%$ & $-4.5\%$,.18 & $+3\%$,.18 \\
\midrule
Frozen L0 s0 & 0.98 & \textbf{$-$0.09} & \textbf{$-$0.34} & \textbf{$-$0.70} & \textbf{$-$0.89} & \textbf{$-$0.90} & \textbf{$-$0.87} & \textbf{$-$0.86} & \textbf{$-$0.82} & +0.00 & \textbf{$-$0.00} & \textbf{$-$0.01} & \textbf{$-$0.90} & \textbf{$-$0.89} \\
Frozen L0 s1 & 0.99 & $-$0.00 & \textbf{$-$0.02} & \textbf{$-$0.17} & \textbf{$-$0.48} & \textbf{$-$0.85} & \textbf{$-$0.93} & \textbf{$-$0.94} & \textbf{$-$0.92} & \textbf{+0.01} & \textbf{$-$0.04} & \textbf{$-$0.09} & \textbf{$-$0.90} & \textbf{$-$0.79} \\
Frozen L0 s2 & 0.98 & $-$0.00 & $-$0.00 & \textbf{$-$0.01} & \textbf{$-$0.08} & \textbf{$-$0.30} & \textbf{$-$0.66} & \textbf{$-$0.85} & \textbf{$-$0.87} & +0.00 & +0.00 & +0.00 & \textbf{$-$0.47} & \textbf{$-$0.20} \\
Frozen L0 s3 & 0.28 & \textbf{$-$0.05} & \textbf{$-$0.16} & \textbf{$-$0.20} & \textbf{$-$0.22} & \textbf{$-$0.21} & \textbf{$-$0.17} & \textbf{$-$0.14} & \textbf{$-$0.11} & \textbf{+0.08} & \textbf{$-$0.06} & \textbf{$-$0.10} & \textbf{$-$0.20} & \textbf{$-$0.21} \\
L0-T s0 & 0.98 & \textbf{$-$0.01} & \textbf{$-$0.02} & \textbf{$-$0.10} & \textbf{$-$0.52} & \textbf{$-$0.83} & \textbf{$-$0.86} & \textbf{$-$0.86} & \textbf{$-$0.84} & +0.00 & $-$0.00 & \textbf{$-$0.01} & \textbf{$-$0.91} & \textbf{$-$0.78} \\
L0-T s1 & 1.00 & $-$0.00 & +0.00 & \textbf{$-$0.03} & \textbf{$-$0.19} & \textbf{$-$0.60} & \textbf{$-$0.86} & \textbf{$-$0.88} & \textbf{$-$0.89} & \textbf{+0.00} & +0.00 & $-$0.00 & \textbf{$-$0.71} & \textbf{$-$0.48} \\
L0-T s2 & 0.99 & $-$0.00 & $-$0.00 & $-$0.00 & \textbf{$-$0.00} & \textbf{$-$0.07} & \textbf{$-$0.38} & \textbf{$-$0.69} & \textbf{$-$0.60} & $-$0.00 & +0.00 & +0.00 & \textbf{$-$0.15} & \textbf{$-$0.03} \\
L0-T s3 & 0.53 & \textbf{$-$0.03} & \textbf{$-$0.12} & \textbf{$-$0.36} & \textbf{$-$0.45} & \textbf{$-$0.45} & \textbf{$-$0.43} & \textbf{$-$0.42} & \textbf{$-$0.41} & \textbf{+0.08} & \textbf{$-$0.08} & \textbf{$-$0.13} & \textbf{$-$0.45} & \textbf{$-$0.46} \\
L1 s0 & 0.96 & \textbf{$-$0.01} & \textbf{$-$0.02} & \textbf{$-$0.03} & \textbf{$-$0.05} & \textbf{$-$0.11} & \textbf{$-$0.41} & \textbf{$-$0.59} & \textbf{$-$0.69} & \textbf{+0.01} & \textbf{$-$0.08} & \textbf{$-$0.15} & \textbf{$-$0.14} & \textbf{$-$0.16} \\
L1 s1 & 1.00 & $-$0.00 & +0.00 & \textbf{$-$0.00} & \textbf{$-$0.06} & \textbf{$-$0.47} & \textbf{$-$0.90} & \textbf{$-$0.94} & \textbf{$-$0.91} & +0.00 & +0.00 & $-$0.00 & \textbf{$-$0.62} & \textbf{$-$0.37} \\
L1 s2 & 1.00 & \textbf{$-$0.00} & \textbf{$-$0.00} & \textbf{$-$0.00} & \textbf{$-$0.00} & \textbf{$-$0.06} & \textbf{$-$0.54} & \textbf{$-$0.67} & \textbf{$-$0.67} & \textbf{$-$0.01} & \textbf{+0.00} & \textbf{+0.00} & \textbf{$-$0.18} & \textbf{$-$0.06} \\
L1 s3 & 0.48 & \textbf{+0.03} & \textbf{$-$0.02} & \textbf{$-$0.18} & \textbf{$-$0.34} & \textbf{$-$0.40} & \textbf{$-$0.38} & \textbf{$-$0.36} & \textbf{$-$0.34} & \textbf{+0.06} & \textbf{$-$0.06} & \textbf{$-$0.10} & \textbf{$-$0.40} & \textbf{$-$0.40} \\
\bottomrule
\end{tabular}
}
\end{table}

\begin{table}[htb]
\centering\scriptsize
\setlength{\tabcolsep}{2.5pt}
\caption{Composition of vehicle outcomes (seed means, Tier~2). VP: valid passage by
lane. The next five columns are the first failure of the vehicle itself or censoring
by another vehicle's reset, as shares of all vehicles; no pass: valid at the horizon
without crossing. World reset: share of worlds reset before the horizon by a
termination other than a map-edge exit. Open on patch: open-lane vehicles with any
wheel contact on the patch.}
\label{tab:decomposition}
\resizebox{\linewidth}{!}{\begin{tabular}{@{}llcccccccccc@{}}
\toprule
Family & $\mu$ & VP closing & VP open & coll. & off-road & incurs. & own term. & censored & no pass & world reset & open on patch \\
\midrule
Frozen L0 & Ref. & 0.797 & 0.819 & 0.016 & 0.002 & 0.000 & 0.004 & 0.168 & 0.000 & 0.254 & 0.011 \\
 & .35 & 0.518 & 0.557 & 0.026 & 0.031 & 0.002 & 0.007 & 0.393 & 0.000 & 0.489 & 0.004 \\
 & .18 & 0.198 & 0.276 & 0.042 & 0.055 & 0.004 & 0.015 & 0.641 & 0.000 & 0.823 & 0.004 \\
 & .08 & 0.025 & 0.178 & 0.092 & 0.056 & 0.021 & 0.019 & 0.697 & 0.000 & 0.951 & 0.009 \\
\addlinespace
L0-T & Ref. & 0.872 & 0.876 & 0.010 & 0.000 & 0.000 & 0.001 & 0.115 & 0.000 & 0.247 & 0.002 \\
 & .35 & 0.731 & 0.769 & 0.032 & 0.020 & 0.000 & 0.001 & 0.194 & 0.000 & 0.260 & 0.001 \\
 & .18 & 0.348 & 0.414 & 0.070 & 0.033 & 0.000 & 0.016 & 0.492 & 0.002 & 0.624 & 0.001 \\
 & .08 & 0.061 & 0.232 & 0.150 & 0.069 & 0.004 & 0.018 & 0.593 & 0.005 & 0.787 & 0.008 \\
\addlinespace
L1 & Ref. & 0.849 & 0.863 & 0.014 & 0.000 & 0.002 & 0.002 & 0.124 & 0.000 & 0.250 & 0.001 \\
 & .35 & 0.785 & 0.817 & 0.022 & 0.012 & 0.002 & 0.003 & 0.158 & 0.000 & 0.251 & 0.000 \\
 & .18 & 0.568 & 0.616 & 0.038 & 0.033 & 0.000 & 0.001 & 0.331 & 0.000 & 0.405 & 0.000 \\
 & .08 & 0.104 & 0.301 & 0.204 & 0.061 & 0.000 & 0.005 & 0.503 & 0.008 & 0.651 & 0.002 \\
\bottomrule
\end{tabular}
}
\end{table}

\begin{table}[htb]
\centering\scriptsize
\caption{Seed-paired family contrasts in valid passage (arms share parent seed
$i$). Reference column: difference in level; other columns: difference in the
change relative to each arm's own reference. Bold: world-bootstrap interval
excludes zero. \Lzerot{} $-$ \Lzero{} includes additional training.}
\label{tab:contrasts}
\begin{tabular}{@{}llccccccccc@{}}
\toprule
Contrast & Seed & Ref. & ..50 & ..35 & ..25 & ..18 & ..12 & ..08 & ..05 & $-8\%$ \\
\midrule
\Lone{} $-$ \Lzerot{} & s0 & \textbf{$-$0.02} & \textbf{+0.01} & \textbf{+0.07} & \textbf{+0.47} & \textbf{+0.72} & \textbf{+0.45} & \textbf{+0.27} & \textbf{+0.15} & \textbf{$-$0.14} \\
 & s1 & \textbf{+0.00} & +0.00 & \textbf{+0.02} & \textbf{+0.13} & \textbf{+0.13} & \textbf{$-$0.04} & \textbf{$-$0.05} & \textbf{$-$0.03} & $-$0.00 \\
 & s2 & \textbf{+0.00} & $-$0.00 & $-$0.00 & $-$0.00 & +0.01 & \textbf{$-$0.16} & +0.02 & \textbf{$-$0.07} & +0.00 \\
 & s3 & \textbf{$-$0.06} & \textbf{+0.10} & \textbf{+0.18} & \textbf{+0.12} & \textbf{+0.05} & \textbf{+0.04} & \textbf{+0.06} & \textbf{+0.07} & \textbf{+0.03} \\
\addlinespace
\Lzerot{} $-$ \Lzero{} & s0 & $-$0.01 & \textbf{+0.32} & \textbf{+0.60} & \textbf{+0.37} & \textbf{+0.07} & +0.01 & +0.00 & \textbf{$-$0.03} & +0.00 \\
 & s1 & +0.00 & \textbf{+0.02} & \textbf{+0.14} & \textbf{+0.29} & \textbf{+0.25} & \textbf{+0.07} & \textbf{+0.05} & \textbf{+0.03} & \textbf{+0.09} \\
 & s2 & \textbf{+0.01} & +0.00 & \textbf{+0.01} & \textbf{+0.08} & \textbf{+0.23} & \textbf{+0.28} & \textbf{+0.15} & \textbf{+0.27} & +0.00 \\
 & s3 & \textbf{+0.25} & \textbf{+0.04} & \textbf{$-$0.16} & \textbf{$-$0.24} & \textbf{$-$0.25} & \textbf{$-$0.26} & \textbf{$-$0.29} & \textbf{$-$0.30} & \textbf{$-$0.03} \\
\bottomrule
\end{tabular}

\end{table}

\begin{table}[htb]
\centering\scriptsize
\caption{Interaction contrasts of grade--friction compositions,
$\mathrm{VP}(g,\mu)-\mathrm{VP}(\mu)-\mathrm{VP}(g)+\mathrm{VP}(\mathrm{ref})$, paired by
world. Negative: the composition loses more than the sum of the single losses. Bold:
world-bootstrap interval excludes zero.}
\label{tab:compound}
\begin{tabular}{@{}lcccc@{}}
\toprule
Checkpoint & $-$4.5\%, $\mu$=0..35 & $-$4.5\%, $\mu$=0..18 & +3.0\%, $\mu$=0..35 & +3.0\%, $\mu$=0..18 \\
\midrule
Frozen L0 s0 & \textbf{$-$0.12} & +0.00 & \textbf{+0.11} & \textbf{+0.01} \\
Frozen L0 s1 & \textbf{$-$0.09} & $-$0.01 & \textbf{+0.05} & \textbf{+0.06} \\
Frozen L0 s2 & \textbf{$-$0.02} & \textbf{$-$0.16} & +0.00 & \textbf{+0.10} \\
Frozen L0 s3 & \textbf{+0.06} & \textbf{+0.07} & \textbf{$-$0.05} & \textbf{$-$0.07} \\
L0-T s0 & \textbf{$-$0.12} & \textbf{$-$0.07} & \textbf{+0.02} & \textbf{+0.05} \\
L0-T s1 & \textbf{$-$0.01} & \textbf{$-$0.11} & +0.01 & \textbf{+0.12} \\
L0-T s2 & +0.00 & \textbf{$-$0.08} & +0.00 & \textbf{+0.04} \\
L0-T s3 & +0.01 & \textbf{+0.09} & $-$0.01 & \textbf{$-$0.07} \\
L1 s0 & \textbf{+0.05} & \textbf{+0.06} & \textbf{$-$0.01} & \textbf{$-$0.05} \\
L1 s1 & \textbf{$-$0.01} & \textbf{$-$0.15} & \textbf{+0.00} & \textbf{+0.09} \\
L1 s2 & +0.00 & \textbf{$-$0.12} & +0.00 & +0.01 \\
L1 s3 & \textbf{+0.05} & \textbf{+0.07} & \textbf{$-$0.02} & \textbf{$-$0.05} \\
\bottomrule
\end{tabular}

\end{table}

\begin{figure}[htb]
\centering
\includegraphics[width=\linewidth]{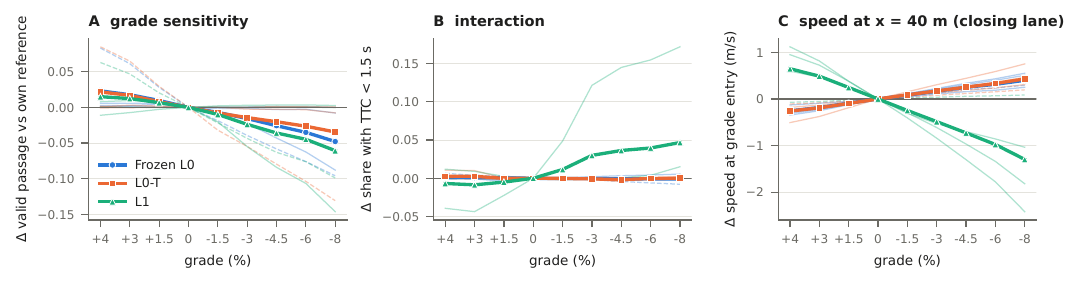}
\caption{Grade sweep, drawn as Figure~\ref{fig:robustness}. (A)~Change in
valid passage. (B)~Change in the share of vehicles with TTC below 1.5~s. (C)~Change in
closing-lane speed where the grade begins; \Lone{} slows before downhill grades.}
\label{fig:grade}
\end{figure}

\paragraph{Hazard analysis.} Table~\ref{tab:hazard} reports the
discrete-time hazard analysis of the 60~Hz subset (64 worlds per cell, seeds pooled
within a family, worlds resampled jointly). A vehicle is at risk in second $b$ while
it is valid, has not crossed and its world has not reset; exposure is the indicator
in the second ending 1~s before $b$, which keeps the terminal pre-contact
divergence of DRAC and TTC out of the failure second. The simpler per-vehicle
question, whether vehicles that ever exceeded their capacity at least 1~s before
their outcome failed more often, gives the opposite sign for several slower
checkpoints (for \Lzerot{} seed~2 at $\muenv=0.08$, 0.19 versus 0.58), because only
vehicles that survive long enough can be exposed. The residual capacity
$a_{\mathrm{avail}}$ contains $\muenv$ by construction, so exceedance frequency
rising as friction falls is not itself a finding; the stratified hazard is.

\begin{table}[htb]
\centering\scriptsize
\setlength{\tabcolsep}{3pt}
\caption{Hazard ratio of a vehicle's own failure given the indicator 1--2~s
earlier, with 95\% world-bootstrap interval and exposed failures / exposed
vehicle-seconds; cells with fewer than three exposed failures show counts only.
Rich subset, seeds pooled.}
\label{tab:hazard}
\resizebox{\linewidth}{!}{\begin{tabular}{@{}llccccc@{}}
\toprule
Indicator (1--2\,s earlier) & Family & $\mu=0$.25 & $\mu=0$.18 & $\mu=0$.12 & $\mu=0$.08 & $\mu=0$.05 \\
\midrule
braking need $>$ residual capacity & Frozen L0 & {\tiny(2/14)} & 29 {\tiny[12, 51]; 12/56} & 29 {\tiny[19, 42]; 25/84} & 21 {\tiny[13, 34]; 24/102} & 18 {\tiny[12, 27]; 33/150} \\
 & L0-T & {\tiny(2/4)} & {\tiny(2/25)} & 21 {\tiny[10, 40]; 14/66} & 15 {\tiny[10, 22]; 34/165} & 11 {\tiny[8, 14]; 53/340} \\
 & L1 & {\tiny(0/3)} & {\tiny(1/11)} & 11 {\tiny[3, 25]; 5/50} & 13 {\tiny[8, 18]; 26/149} & 11 {\tiny[8, 15]; 36/212} \\
\addlinespace
TTC $<1.5$\,s & Frozen L0 & 26 {\tiny[14, 44]; 21/174} & 25 {\tiny[16, 36]; 33/193} & 36 {\tiny[29, 45]; 99/310} & 30 {\tiny[23, 38]; 89/303} & 21 {\tiny[16, 27]; 69/294} \\
 & L0-T & 30 {\tiny[21, 40]; 65/687} & 37 {\tiny[28, 50]; 82/452} & 29 {\tiny[23, 36]; 113/473} & 24 {\tiny[20, 28]; 167/634} & 19 {\tiny[16, 23]; 175/744} \\
 & L1 & 39 {\tiny[18, 70]; 27/367} & 15 {\tiny[7, 26]; 22/444} & 17 {\tiny[12, 23]; 68/510} & 22 {\tiny[18, 27]; 175/700} & 19 {\tiny[16, 23]; 184/740} \\
\addlinespace
capacity exceeded, TTC $\ge1.5$\,s & Frozen L0 & {\tiny(1/11)} & 24 {\tiny[6, 47]; 7/43} & 25 {\tiny[12, 43]; 13/61} & 16 {\tiny[8, 31]; 12/78} & 22 {\tiny[13, 33]; 27/115} \\
 & L0-T & {\tiny(0/1)} & {\tiny(2/25)} & 14 {\tiny[5, 30]; 6/54} & 10 {\tiny[5, 15]; 14/136} & 10 {\tiny[6, 13]; 31/272} \\
 & L1 & {\tiny(0/2)} & {\tiny(0/10)} & {\tiny(2/42)} & 12 {\tiny[7, 18]; 17/128} & 10 {\tiny[6, 14]; 21/167} \\
\bottomrule
\end{tabular}
}
\end{table}

\begin{figure}[htb]
\centering
\includegraphics[width=\linewidth]{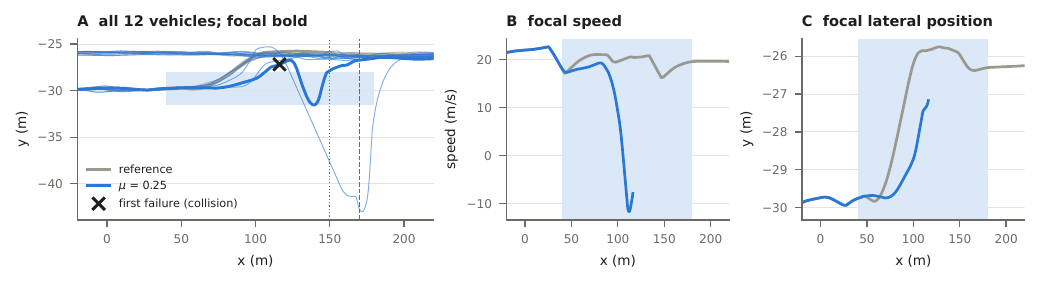}
\caption{One matched pair (\Lzero{} seed~0, scenario 100050): reference
road (grey) and $\muenv=0.25$ patch (blue, shaded). (A)~All 12 vehicles; focal vehicle
bold, first failure marked. (B, C)~Focal vehicle's speed and lateral position until its
failure.}
\label{fig:pair}
\end{figure}

\paragraph{Illustrative matched pair.} Figure~\ref{fig:pair} shows one matched
pair selected by a rule fixed before plotting: \Lzero{} seed~0, reference versus
$\muenv=0.25$; the first world of the rich subset, in scenario-seed order, in which
neither world resets, a closing-lane vehicle validly passes on the reference road
and the same vehicle's first failure on the patch is a collision, off-road
departure or incursion. This is world 100050; it illustrates one mechanism and is
not representative of the population.

\begin{table}[htb]
\centering\scriptsize
\setlength{\tabcolsep}{2.5pt}
\caption{Tier-1 valid passage (32 worlds) on the reference road and its change for
selected conditions. Bold: the 95\% paired world-bootstrap interval excludes zero.
Learned rows are the first 32 worlds of the Tier-2 cells.}
\label{tab:tier1}
\resizebox{\linewidth}{!}{\begin{tabular}{@{}lcccccccccccccc@{}}
\toprule
Checkpoint & Ref. & ..70 & ..50 & ..35 & ..25 & ..18 & ..12 & ..08 & ..05 & $+4\%$ & $-4.5\%$ & $-8\%$ & $-4.5\%$,.18 & $+3\%$,.18 \\
\midrule
IDM-17 & 0.68 & +0.01 & +0.01 & $-$0.01 & $-$0.04 & \textbf{$-$0.18} & \textbf{$-$0.31} & \textbf{$-$0.43} & \textbf{$-$0.49} & \textbf{$-$0.08} & +0.04 & \textbf{+0.05} & \textbf{$-$0.20} & \textbf{$-$0.18} \\
PDM-Closed-17 & 1.00 & +0.00 & +0.00 & \textbf{$-$0.06} & \textbf{$-$0.16} & \textbf{$-$0.19} & \textbf{$-$0.32} & \textbf{$-$0.46} & \textbf{$-$0.48} & \textbf{$-$0.04} & $-$0.01 & $-$0.01 & \textbf{$-$0.36} & \textbf{$-$0.14} \\
Frozen L0 s0 & 0.98 & \textbf{$-$0.10} & \textbf{$-$0.36} & \textbf{$-$0.62} & \textbf{$-$0.84} & \textbf{$-$0.91} & \textbf{$-$0.89} & \textbf{$-$0.84} & \textbf{$-$0.81} & +0.01 & $-$0.01 & $-$0.01 & \textbf{$-$0.91} & \textbf{$-$0.90} \\
Frozen L0 s1 & 0.99 & +0.00 & $-$0.03 & $-$0.06 & \textbf{$-$0.46} & \textbf{$-$0.86} & \textbf{$-$0.90} & \textbf{$-$0.96} & \textbf{$-$0.92} & +0.00 & $-$0.01 & \textbf{$-$0.08} & \textbf{$-$0.88} & \textbf{$-$0.80} \\
Frozen L0 s2 & 0.98 & $-$0.02 & +0.01 & $-$0.03 & $-$0.06 & \textbf{$-$0.24} & \textbf{$-$0.64} & \textbf{$-$0.84} & \textbf{$-$0.87} & $-$0.01 & +0.02 & \textbf{+0.02} & \textbf{$-$0.42} & \textbf{$-$0.15} \\
Frozen L0 s3 & 0.26 & $-$0.03 & \textbf{$-$0.13} & \textbf{$-$0.18} & \textbf{$-$0.20} & \textbf{$-$0.18} & \textbf{$-$0.14} & \textbf{$-$0.12} & \textbf{$-$0.08} & \textbf{+0.11} & \textbf{$-$0.05} & \textbf{$-$0.10} & \textbf{$-$0.16} & \textbf{$-$0.20} \\
L0-T s0 & 0.96 & $-$0.00 & $-$0.01 & \textbf{$-$0.08} & \textbf{$-$0.48} & \textbf{$-$0.77} & \textbf{$-$0.86} & \textbf{$-$0.87} & \textbf{$-$0.82} & +0.01 & $-$0.01 & $-$0.01 & \textbf{$-$0.89} & \textbf{$-$0.75} \\
L0-T s1 & 0.98 & +0.00 & +0.00 & +0.00 & \textbf{$-$0.17} & \textbf{$-$0.42} & \textbf{$-$0.84} & \textbf{$-$0.86} & \textbf{$-$0.87} & +0.00 & +0.00 & +0.00 & \textbf{$-$0.63} & \textbf{$-$0.35} \\
L0-T s2 & 1.00 & +0.00 & $-$0.01 & +0.00 & $-$0.02 & $-$0.02 & \textbf{$-$0.37} & \textbf{$-$0.73} & \textbf{$-$0.61} & $-$0.01 & $-$0.01 & $-$0.01 & \textbf{$-$0.16} & \textbf{$-$0.10} \\
L0-T s3 & 0.47 & $-$0.01 & \textbf{$-$0.12} & \textbf{$-$0.29} & \textbf{$-$0.36} & \textbf{$-$0.39} & \textbf{$-$0.37} & \textbf{$-$0.36} & \textbf{$-$0.36} & \textbf{+0.10} & \textbf{$-$0.08} & \textbf{$-$0.11} & \textbf{$-$0.38} & \textbf{$-$0.39} \\
L1 s0 & 0.93 & +0.00 & $-$0.01 & $-$0.03 & $-$0.02 & \textbf{$-$0.05} & \textbf{$-$0.36} & \textbf{$-$0.61} & \textbf{$-$0.64} & $-$0.01 & \textbf{$-$0.05} & \textbf{$-$0.14} & $-$0.05 & \textbf{$-$0.12} \\
L1 s1 & 1.00 & +0.00 & +0.00 & +0.00 & $-$0.06 & \textbf{$-$0.49} & \textbf{$-$0.93} & \textbf{$-$0.95} & \textbf{$-$0.91} & +0.00 & +0.00 & +0.00 & \textbf{$-$0.59} & \textbf{$-$0.45} \\
L1 s2 & 1.00 & +0.00 & $-$0.01 & $-$0.01 & $-$0.01 & $-$0.04 & \textbf{$-$0.35} & \textbf{$-$0.71} & \textbf{$-$0.65} & $-$0.02 & +0.00 & +0.00 & \textbf{$-$0.27} & $-$0.04 \\
L1 s3 & 0.44 & +0.02 & $-$0.01 & \textbf{$-$0.11} & \textbf{$-$0.20} & \textbf{$-$0.37} & \textbf{$-$0.36} & \textbf{$-$0.32} & \textbf{$-$0.30} & \textbf{+0.07} & \textbf{$-$0.06} & \textbf{$-$0.09} & \textbf{$-$0.36} & \textbf{$-$0.38} \\
\bottomrule
\end{tabular}
}
\end{table}

\paragraph{Release materials.} The release includes the analysis scripts, vehicle-
and world-level tables, per-checkpoint and per-seed estimates, bootstrap outputs
and the data behind every figure and table.

\end{document}